%% file: main.tex
\documentclass[]{infiniai}

\usepackage{microtype}
\usepackage{amsfonts}
\usepackage{graphicx}
\usepackage{booktabs}
\usepackage{wrapfig}
\usepackage{amsmath}
\usepackage{adjustbox}
\usepackage{amsthm}
\usepackage{subcaption}
\usepackage{appendix}
\usepackage{algpseudocode}
\usepackage[linesnumbered,lined,boxed,commentsnumbered,ruled,longend]{algorithm2e}
\usepackage[capitalize,noabbrev]{cleveref}
\theoremstyle{plain}
\newtheorem{theorem}{Theorem}[section]

\theoremstyle{definition}

\theoremstyle{remark}

\usepackage{enumitem}
\DeclareMathOperator{\softmax}{softmax}
\DeclareMathOperator*{\argmax}{arg\,max}
\usepackage{xspace}
\newcommand{\Sys}{\textsc{MonarchRT}\xspace}

\usepackage{subcaption}
\usepackage{tabularx}
\usepackage{array}
\usepackage{multirow}
\usepackage{makecell}
\usepackage{minted}
\usepackage[T1]{fontenc}
\usepackage{textcomp}   
\usepackage{lmodern}    
\newcolumntype{Y}{>{\centering\arraybackslash}X}
\usepackage{float}

\definecolor{lavenderpink}{rgb}{0.95, 0.85, 0.95}
\definecolor{softlavender}{rgb}{0.64, 0.50, 0.68}
\title{MonarchRT: Efficient Attention for Real-Time Video Generation}

\author{Krish Agarwal$^1$}
\author{Zhuoming Chen$^1$}
\author{Cheng Luo}
\author{Yongqi Chen$^3$}
\author{Haizhong Zheng$^1$}
\author{Xun Huang$^3$}
\author{Atri Rudra$^2$}
\author{Beidi Chen$^1$}
\affiliation{$^1$Carnegie Mellon University\\$^2$University at Buffalo\\$^3$Morpheus AI}
\contact{\{krisha2,zhuominc,haizhonz,beidic\}@andrew.cmu.edu\\wdlctc@gmail.com, yongqichcd@gmail.com, xunhuang1995@gmail.com, atri@buffalo.edu}
\abstract{
Real-time video generation with Diffusion Transformers is bottlenecked by the quadratic cost of 3D self-attention, especially in real-time regimes that are both few-step and autoregressive, where errors compound across time and each denoising step must carry substantially more information. In this setting, we find that prior sparse-attention approximations break down, despite showing strong results for bidirectional, many-step diffusion. Specifically, we observe that video attention is not reliably sparse, but instead combines pronounced periodic structure driven by spatiotemporal position with dynamic, sparse semantic correspondences and dense mixing, exceeding the representational capacity of even oracle top-$k$ attention. Building on this insight, we propose \textbf{\Sys}, a structured attention parameterization for video diffusion models that factorizes attention using Monarch matrices. Through appropriately aligned block structure and our extended \emph{tiled Monarch parameterization}, we achieve high expressivity while preserving computational efficiency. We further overcome the overhead of parameterization through finetuning, with custom Triton kernels. We first validate the high efficacy of \Sys over existing sparse baselines designed only for bidirectional models. We further observe that \Sys attains up to \emph{95\% attention sparsity} with no loss in quality when applied to the state-of-the-art model Self-Forcing, making Monarch-RT a pioneering work on highly-capable sparse attention parameterization for real-time video generation. Our optimized implementation outperforms FlashAttention-2, FlashAttention-3, and FlashAttention-4 kernels on Nvidia RTX 5090, H100, and B200 GPUs respectively, providing kernel speedups in the range of 1.4-11.8$\times$. This enables us, for the first time, to achieve true real-time video generation with Self-Forcing at 16 FPS on a single RTX 5090.
}
\metadata[Github]{\url{https://github.com/Infini-AI-Lab/MonarchRT}}
\metadata[Website]{\url{https://infini-ai-lab.github.io/MonarchRT}}

\begin{document}

\maketitle
\input{sections/introduction}
\input{sections/background}
\input{sections/analysis}
\input{sections/method}

\input{sections/experiment}
\input{sections/discussion}
\clearpage
\newpage
\bibliographystyle{assets/plainnat}
\bibliography{paper}
\clearpage
\newpage
\beginappendix
\input{sections/supp}
\end{document}

%% file: sections/introduction.tex
\section{Introduction}

\begin{figure*}[t]
  \centering
  \subfloat[
    \label{fig:motivating_sparsity_error_comparison}
  ]{
    \includegraphics[width=0.34\linewidth]{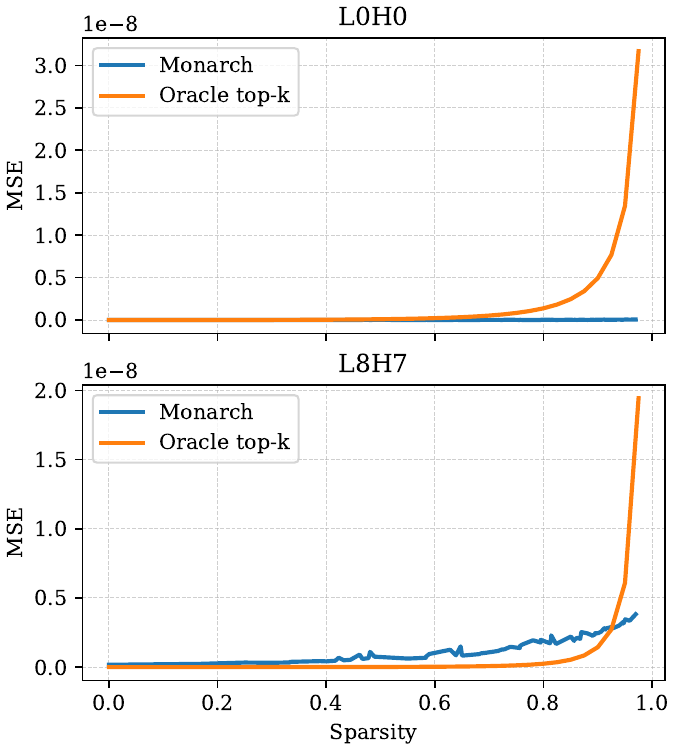}
  }
  \hfill
  \subfloat[
    \label{fig:motivating_example_generations}
  ]{
    \includegraphics[width=0.6\linewidth]{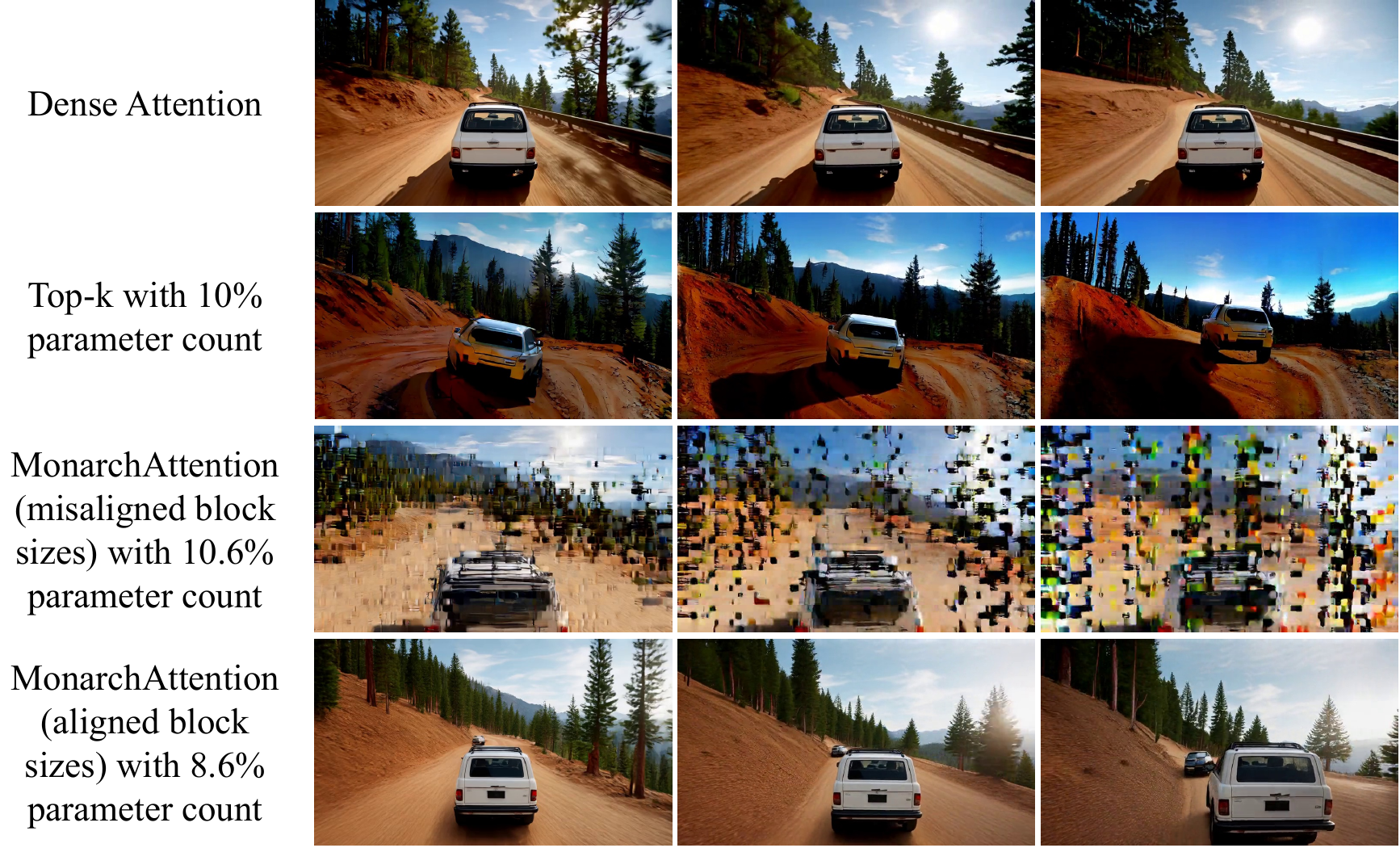}
  }

  \caption{\textbf{Left:} MSE of oracle top-$k$ and Monarch parameterizations of an attention map compared to the original dense attention map for varying levels of sparsity. Results shown for two different layers/heads on Self-Forcing. Monarch incurs much lower error for high levels of sparsity. \textbf{Right:} Example generations on the same prompt on Self-Forcing. First row shows exact top-$k$ with 10\% density still produces poor quality output. Third row shows that using inference-only MonarchAttention (with aligned block sizes) produces higher output quality with a lower parameter count of 8.6\%. Second row shows that, although the parameter count is increased to 10.6\%, inference-only MonarchAttention with misaligned block sizes incurs pixel-level permutation effects. Parameter count refers to the number of parameters used to estimate the full attention map.}
  \label{fig:motivating_example}
\end{figure*}

\begin{figure*}[t]
  \centering
  \includegraphics[width=\linewidth]{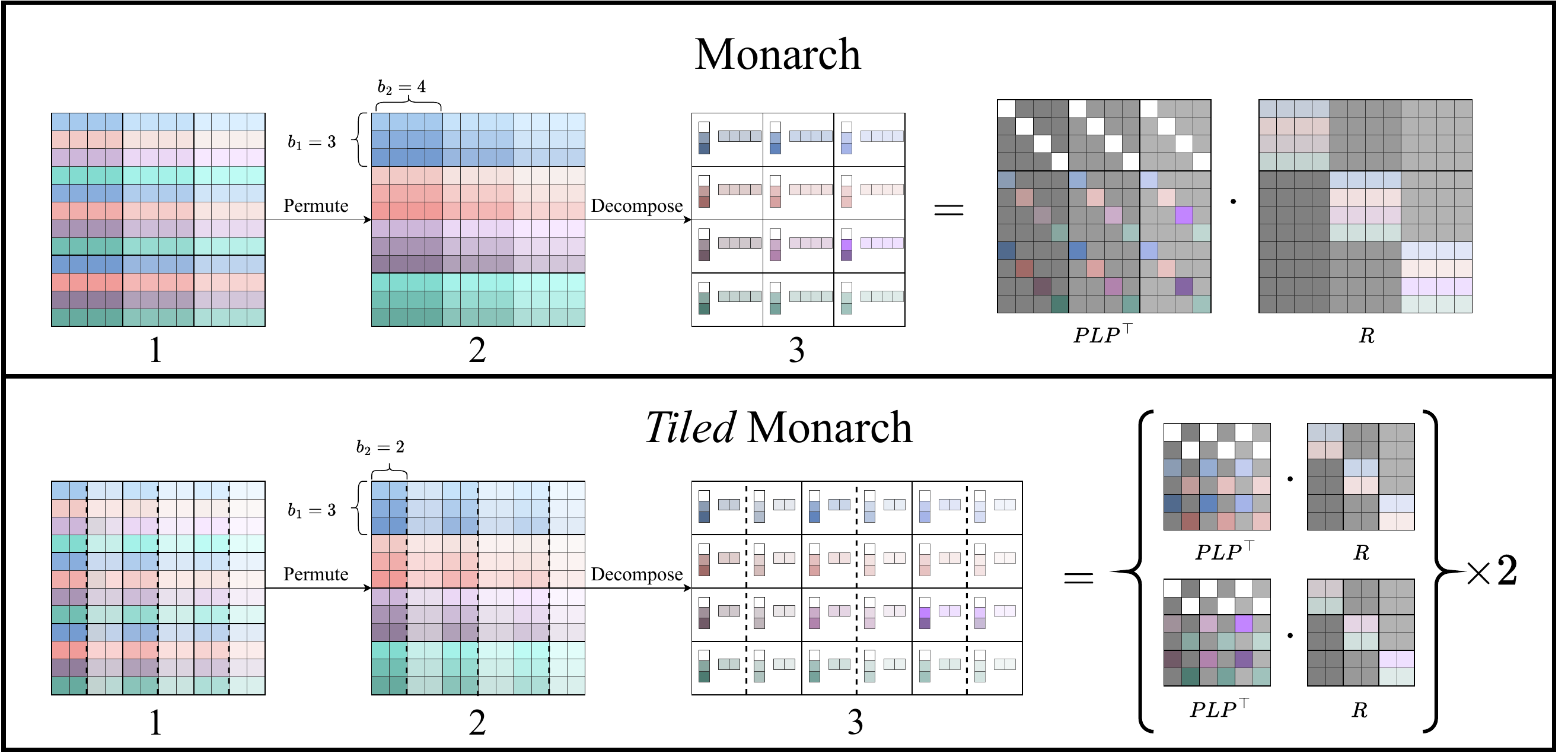}
 \caption{\textbf{Illustration of Regular and Tiled Monarch Parameterization.}
Top: An example of Monarch parameterization applied to a $12 \times 12$ matrix with block size
$(b_1, b_2) = (3, 4)$. Bottom: An example of \emph{tiled} Monarch parameterization applied with block size
$(b_1, b_2) = (3, 2)$.
\textbf{(1)} The original matrix is first permuted to expose an implicit block-wise low-rank
structure.
\textbf{(2)} After permutation, the matrix is reorganized into blocks of size $b_1 \times b_2$,
where each block corresponds to a group of rows and columns.
\textbf{(3)} Each block is then independently decomposed into low-rank factors.
Overall, Monarch represents the matrix as $PLP^{\top}R$, where $P$ denotes a permutation
matrix, $PLP^{\top}$ is a block-wise diagonal matrix, and $R$ is a block-diagonal matrix. The tiled Monarch parameterization has 2$\times$ the parameter count and is strictly more expressive than the regular Monarch parameterization.
}
  \label{fig:monarch_parameterization}
\end{figure*}

With the common understanding that substantial redundancy exists in 3D attention, many approximation algorithms have been proposed to reduce its computational cost. However, we found that two key properties critical to real-time video generation~\citep{zhang2025turbodiffusion}, namely \textbf{auto-regressiveness}~\citep{huang2025selfforcingbridgingtraintest,teng2025magi,yin2025slow} and the use of \textbf{fewer diffusion steps}~\citep{yin2024onestepdiffusiondistributionmatching,yin2024improveddistributionmatchingdistillation,liu2024distilled,liu2025distilled}, can significantly amplify approximation difficulties. Auto-regressive generation accumulates errors over time, while shortening the diffusion process compresses computation, causing each diffusion step to process substantially more information. Critically, these features are key in the design of real-time, \emph{interactive} video generation models, including Genie 3~\citep{genie3}, WorldPlay~\citep{sun2025worldplaylongtermgeometricconsistency}, and LingBot-World~\citep{robbyantteam2026advancingopensourceworldmodels}.

Prior efficient approximations for 3D attention are fundamentally constrained by their
\textbf{limited expressiveness}, which leads to substantial degradation in generated video quality.
\emph{Sparse attention} methods capture \textbf{positional patterns} by restricting attention to
temporally and spatially local neighborhoods~\citep{li2025radialattentiononlogn,
xi2025sparsevideogenacceleratingvideo, zhang2025fastvideogenerationsliding},
or capture \textbf{semantic patterns} by attending to a small subset of tokens selected via
clustering or retrieval-based strategies~\citep{zhang2025vsafastervideodiffusion,
yang2025sparse, zhang2025slasparsitydiffusiontransformers}.
The former class weakens the inherent ability of DiTs to model long-range semantic dependencies,
effectively regressing toward convolution-based architectures~\citep{10.1007/978-3-319-46723-8_49,
ronneberger2015unetconvolutionalnetworksbiomedical}, whose representational limitations are well known.
The latter class, while more flexible, is highly parameter-inefficient when
\textbf{dense mixing} is required, as evidenced by the failure of oracle top-$k$ attention even with
$10\%$ of the FLOPs of full attention (\Cref{fig:motivating_example_generations}). \emph{Low-rank and linear attention} methods similarly struggle to represent long-range semantic
patterns empirically~\citep{zhou2025gsminfinitellmsbehaveinfinitely,
he2025alleviatingforgetfulnesslinearattention} and often
requiring substantial retraining to adapt from DiTs~\citep{zhang2025slasparsitydiffusiontransformers}.
As a result, to preserve video quality, existing approaches typically retain more than $40\%$ of the
FLOPs of full attention~\citep{zhang2025slasparsitydiffusiontransformers, yang2025sparse,
xi2025sparsevideogenacceleratingvideo}.
While such compromises can be effective for many-step diffusion models such as Wan~2.1~\citep{wan2025wanopenadvancedlargescale},
which exploit redundancy across approximately $50$ diffusion steps, they are fundamentally
incompatible with state-of-the-art real-time video generation systems that operate with far fewer
steps~\citep{huang2025selfforcingbridgingtraintest, zhang2025turbodiffusion}.

Therefore, to preserve the strengths of DiTs while remaining viable for real-time video generation,
an ideal attention approximation must simultaneously capture three critical patterns:
\textbf{positional patterns}, \textbf{semantic patterns}, and \textbf{dense mixing}.
This naturally raises the question of whether there exists an approach that is substantially more
computationally efficient than full attention while retaining strong expressive power.

Beyond sparse and low-rank methodologies, Monarch~\citep{dao2022monarchexpressivestructuredmatrices,
dao2021kaleidoscopeefficientlearnablerepresentation, desa2017prongedprogressstructureddense}
introduces a unifying class of structured matrices that can efficiently represent a broad family of
linear operators, including sparse and low-rank matrices as well as structured transforms such as
FFT and Hadamard.
More recently, MonarchAttention~\citep{yaras2025monarchattentionzeroshotconversionfast}, which employs an iterative optimization procedure to efficiently obtain Monarch factors, demonstrates
that attention matrices themselves can be effectively approximated using Monarch parameterizations.
Our empirical analysis (\Cref{fig:motivating_sparsity_error_comparison}) further reveals that Monarch parameterizations recover 3D attention matrices
significantly better than oracle top-$k$ or low-rank approximations, pointing to a promising
direction for accelerating real-time video generation without sacrificing expressiveness.

However, leveraging Monarch parameterizations to approximate 3D attention in practical
real-time video generation systems introduces several non-trivial technical challenges. \underline{First} (\textbf{Shape Alignment}),
unlike Monarch in MLPs~\citep{dao2022monarchexpressivestructuredmatrices, fu2023monarch}, where the
semantics of input and output channels can be learned implicitly during training, the rows and
columns of 3D attention matrices are explicitly tied to physical pixel patches in a video.
Consequently, when the structural assumptions of Monarch parameterizations (e.g., block sizes) are
misaligned with the spatial--temporal layout of pixel patches, the approximation quality can
degrade dramatically, often leading to severe artifacts or even complete collapse of the generated
videos. \underline{Second} (\textbf{Limited Flexibility}),
although Monarch allows adjusting computational cost by varying block sizes, we observe a
systematic issue in which the approximation error does not reliably decrease as more FLOPs are
allocated.
This lack of monotonic refinement fundamentally limits Monarch's ability to progressively improve
approximation quality, in sharp contrast to sparse-attention-based methods whose accuracy typically
scales with increased computation. \underline{Third} (\textbf{High Runtime Overhead}),
the iterative refinement strategy adopted by MonarchAttention~\citep{yaras2025monarchattentionzeroshotconversionfast}
requires multiple refinement steps to achieve high-quality approximations, rendering it
computationally prohibitive for real-time video generation systems with strict latency constraints.

To address these challenges, we introduce \Sys, a unified framework for applying Monarch
parameterizations to video generation.
\Sys first provides a factorization structure that is explicitly compatible with 3D attention,
and further enables \emph{arbitrarily accurate} matrix approximations through our proposed
\emph{Tiled Monarch Parameterizations} (illustrated in~\Cref{fig:monarch_parameterization} Bottom).
In addition, by finetuning MonarchAttention, we substantially reduce the number of iterative
refinement steps required during inference.
These components make Monarch parameterization both accurate and efficient for
real-time video generation.

\noindent In summary, our contributions are as follows:
\begin{itemize}
    \item In~\Cref{sec:topkfail,sec:attentionmodel,sec:monarchfit}, we analyze the structural
    patterns of 3D video attention, explaining why existing sparse- and low-rank-based methods fail
    to capture them, and why Monarch provides a principled and expressive parameterization for 3D
    attention.

    \item In~\Cref{sec:challenges}, we identify the key practical challenges in applying Monarch
    parameterizations to video models, including \emph{shape misalignment}, \emph{limited
    flexibility}, and \emph{parameterization overhead}.

    \item In~\Cref{sec:grouping,sec:tiled}, we introduce \emph{Tiled Monarch Parameterization},
    together with a simple yet crucial block-alignment strategy.
    This formulation generalizes the original Monarch parameterization and enables arbitrarily
    accurate approximations of 3D attention matrices.

    \item In~\Cref{sec:implementation}, we demonstrate how training can be leveraged to minimize
    inference-time cost by reducing the number of iterative refinement steps.
    We further provide an efficient Triton implementation that supports both forward and backward
    passes of Monarch attention.
\end{itemize}

In~\Cref{sec:experiments}, we conduct comprehensive evaluations of \Sys across multiple video
generation settings.
We benchmark our method against existing sparse-attention-based approaches on
auto-regressive video models (e.g., Self-Forcing), demonstrating that \Sys achieves the highest
visual quality under real-time generation constraints.
While prior sparse attention methods fail to maintain visual fidelity beyond $85\%$ sparsity,
\Sys preserves high generation quality even when reducing attention computation by $95\%$. We further apply \Sys to bidirectional video diffusion models (e.g., Wan 2.1-1.3B ~\citep{wan2025wanopenadvancedlargescale})
and compare it against additional sparse/dense baselines.
\Sys consistently achieves comparable generation quality while significantly reducing
computational cost. Using our optimized kernel implementation, \Sys obtains up to a 5.6$\times$ speedup over FA-3~\citep{shah2024flashattention3fastaccurateattention}
on H100 and 11.8$\times$ over FA-2~\citep{dao2022flashattentionfastmemoryefficientexact,dao2023flashattention}
on RTX 5090. Notably, with \Sys we are able to achieve true real-time generation at 16 FPS with Self-Forcing on RTX 5090.

%% file: sections/background.tex
\section{Background}

\label{sec:background}
In this section, we first discuss related work on Monarch parameterizations and MonarchAttention. Then we provide a visualization to illustrate the Monarch parameterization process. Finally, we formally present the attention approximation problem.

\subsection{Monarch}

\paragraph{Monarch Parameterization.}
Monarch parameterization~\citep{dao2022monarchexpressivestructuredmatrices} represents an
$N \times N$ matrix $\boldsymbol{M}$, with $N = b_1 b_2$, using a structured factorization defined
by two block sizes $b_1$ and $b_2$.
Concretely, $\boldsymbol{M}$ is expressed as
\[
\boldsymbol{M} = \boldsymbol{P} \boldsymbol{L} \boldsymbol{P}^{\top} \boldsymbol{R},
\]
where $\boldsymbol{P}$ is a fixed permutation matrix that reshapes a length-$N$ vector into a
$b_1 \times b_2$ matrix, transposes it, and flattens it back.
This permutation exposes an implicit block-wise structure in $\boldsymbol{M}$, as illustrated
in~\Cref{fig:monarch_parameterization}. After permutation, $\boldsymbol{L}$ is block-diagonal with $b_2$ blocks of size
$b_1 \times b_1$, each operating independently on a group of rows, while $\boldsymbol{R}$ is
block-diagonal with $b_1$ blocks of size $b_2 \times b_2$, each operating on a group of columns.
Intuitively, Monarch assumes that, after permutation, the matrix can be decomposed into
independent low-rank blocks along the two axes. Equivalently, Monarch can be interpreted as a block-wise rank-$1$ structure under permutation.
Specifically, $\boldsymbol{M}$ can be viewed as a $4$D tensor of shape
$b_1 \times b_2 \times b_1 \times b_2$, with entries defined as
\[
\boldsymbol{M}_{\ell j k i} = \boldsymbol{L}_{j \ell k} \, \boldsymbol{R}_{k j i}.
\]
This perspective leads to an efficient projection algorithm: given an arbitrary $N \times N$
matrix, one first permutes it, reshapes it into the $4$D form, and then performs independent
rank-$1$ projections on each post-permutation $2$D slice (along the $\ell$ and $i$ dimensions) to
populate $\boldsymbol{L}$ and $\boldsymbol{R}$. The class of Monarch matrices strictly generalizes butterfly matrices.
Since sparse matrices can be represented as products of butterfly matrices~\citep{dao2021kaleidoscopeefficientlearnablerepresentation},
the same representational guarantees extend to Monarch parameterizations.
An example with $N=12$ and $(b_1,b_2)=(3,4)$ is visualized in~\Cref{fig:monarch_parameterization} Top.

\begin{figure*}[t]
  \centering
  \includegraphics[width=\linewidth]{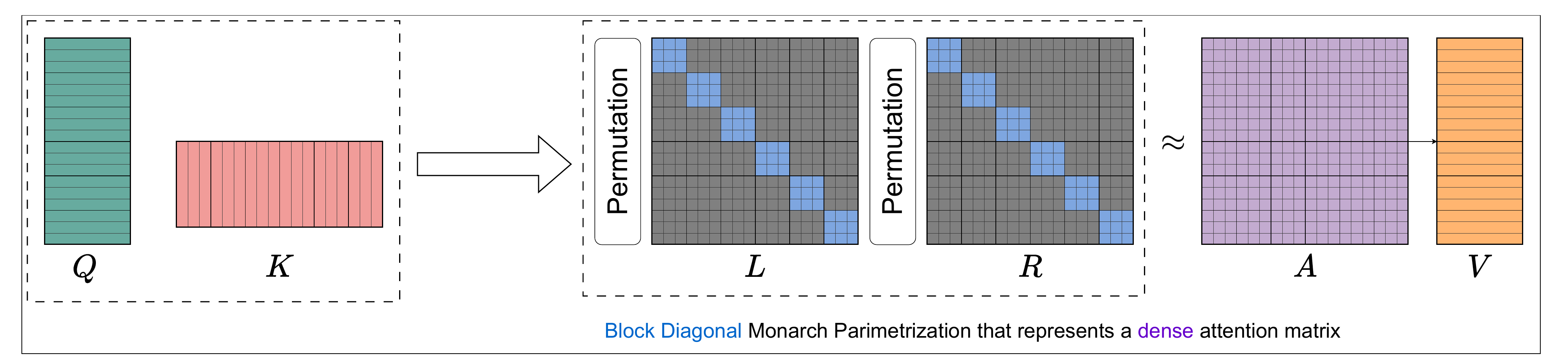}
  \caption{\textbf{Overview of the MonarchAttention pipeline.}
Given query and key matrices $(\boldsymbol{Q}, \boldsymbol{K})$, MonarchAttention iteratively
refines the Monarch factors $\boldsymbol{L}$ and $\boldsymbol{R}$, each composed of sparse
block-diagonal matrices.
At each iteration, one factor is updated while the other is held fixed, without explicitly
materializing the full attention matrix.
Despite the highly structured and sparse parameterization of $\boldsymbol{L}$ and $\boldsymbol{R}$,
the resulting attention matrix
$\boldsymbol{A} \approx \boldsymbol{P}\boldsymbol{L}\boldsymbol{P}^{\top}\boldsymbol{R}$
is dense, highlighting the strong expressiveness of Monarch parameterization.
Algorithmic details are provided in Appendix~\ref{app:algodetails}.}
  \label{fig:monarch_attention}
\end{figure*}

\paragraph{MonarchAttention}
aims to approximate the full attention matrix
$\boldsymbol{A} = \softmax(\boldsymbol{Q}\boldsymbol{K}^{\top}) \in \mathbb{R}^{N \times N}$
using Monarch parameterization with factors
$\boldsymbol{L} \in \mathbb{R}^{b_2 \times b_1 \times b_1}$ and
$\boldsymbol{R} \in \mathbb{R}^{b_1 \times b_2 \times b_2}$, where block sizes
$b_1$ and $b_2$ satisfy $N = b_1 b_2$~\citep{yaras2025monarchattentionzeroshotconversionfast}.
A direct approach would require explicitly forming $\boldsymbol{A}$ and projecting it onto the
Monarch structure via SVD on each permuted block, which is computationally prohibitive. To avoid materializing the full attention matrix, MonarchAttention proposes an iterative refinement
algorithm that directly optimizes the Monarch factors.
Leveraging an alternative variational formulation of softmax~\citep{blondel2019learningclassifiersfenchelyounglosses},
the attention matrix can be expressed as
\[
\boldsymbol{A}
= \argmax_{\boldsymbol{A}_i \in \Delta^N}
\langle \boldsymbol{A}, \boldsymbol{Q}\boldsymbol{K}^{\top} \rangle + H(\boldsymbol{A}),
\]
where $H(\cdot)$ denotes the entropy. By constraining $\boldsymbol{A}$ to lie in the Monarch family, it can be interpreted as a
$b_1 \times b_2 \times b_1 \times b_2$ tensor with entries
$\boldsymbol{A}_{\ell j k i} = \boldsymbol{L}_{j \ell k}\boldsymbol{R}_{k j i}$.
This formulation allows the objective to be optimized directly over $\boldsymbol{L}$ and
$\boldsymbol{R}$.
Through careful manipulation of the objective and by imposing slightly stronger constraints on the
factors, MonarchAttention enables alternating maximization: updating $\boldsymbol{R}$ while holding
$\boldsymbol{L}$ fixed, and vice versa. After a reasonable number of iterations, the algorithm produces Monarch factors that can be applied
sequentially to the value matrix $\boldsymbol{V}$ to compute the attention output, without ever
constructing the dense attention matrix explicitly. The pipeline is illustrated in~\Cref{fig:monarch_attention}.
Additional algorithmic details are provided in~\Cref{app:algodetails}.

\subsection{Problem Formulation}
Given query, key, and value matrices
$\boldsymbol{Q}, \boldsymbol{K}, \boldsymbol{V} \in \mathbb{R}^{n \times d}$,
the attention output is
\[
\mathrm{Attn}(\boldsymbol{Q}, \boldsymbol{K}, \boldsymbol{V})
= \softmax(\boldsymbol{Q}\boldsymbol{K}^{\top}) \boldsymbol{V},
\]
where the softmax is applied row-wise.
We denote the attention matrix by
$\boldsymbol{A} = \softmax(\boldsymbol{Q}\boldsymbol{K}^{\top}) \in \mathbb{R}^{n \times n}$.
The goal of attention parameterization is to reduce the quadratic cost of computing
$\boldsymbol{A}$ while preserving approximation quality.

Formally, given an approximation procedure $f(\boldsymbol{Q}, \boldsymbol{K})$, we seek to balance
the approximation error
\[
\mathbb{E}\!\left[ \bigl\| f(\boldsymbol{Q}, \boldsymbol{K}) - \boldsymbol{A} \bigr\|_F^2 \right],
\]
and the computational cost $\mathcal{C}\!\left(f(\cdot)\right)$.

We consider three representative parameterization families, each imposing a distinct structural
constraint on $\boldsymbol{A}$:
\begin{enumerate}
  \item \textbf{Sparse.}
  \[
    \boldsymbol{A}_{\mathrm{s}}
    = \arg\min_{\boldsymbol{A}' \in \mathbb{R}^{n \times n} \,:\, \mathrm{NNZ}(\boldsymbol{A}')}
    \| \boldsymbol{A}' - \boldsymbol{A} \|_{F}^{2}.
  \]

  \item \textbf{Low-rank.}
  \[
    \boldsymbol{A}_{\ell}
    = \arg\min_{\boldsymbol{\bar{Q}},\, \boldsymbol{\bar{K}} \in \mathbb{R}^{n \times \bar{n}}}
    \| \boldsymbol{\bar{Q}} \boldsymbol{\bar{K}}^{\top} - \boldsymbol{A} \|_{F}^{2},
    \quad \bar{n} < n.
  \]

  \item \textbf{Monarch.}
  \[
    \boldsymbol{A}_{\mathrm{m}}
    = \arg\min_{\boldsymbol{A}' \in \mathcal{M}}
    \| \boldsymbol{A}' - \boldsymbol{A} \|_{F}^{2},
  \]
  where $\mathcal{M}$ denotes the family of Monarch matrices.
\end{enumerate}

An ideal parameterization achieves low approximation error with minimal computational cost,
which is ultimately governed by the structural properties of $\boldsymbol{A}$.

%% file: sections/analysis.tex
\begin{figure*}[t]
  \centering
  \subfloat[
    \label{fig:attnscores}
  ]{
    \includegraphics[width=0.34\linewidth]{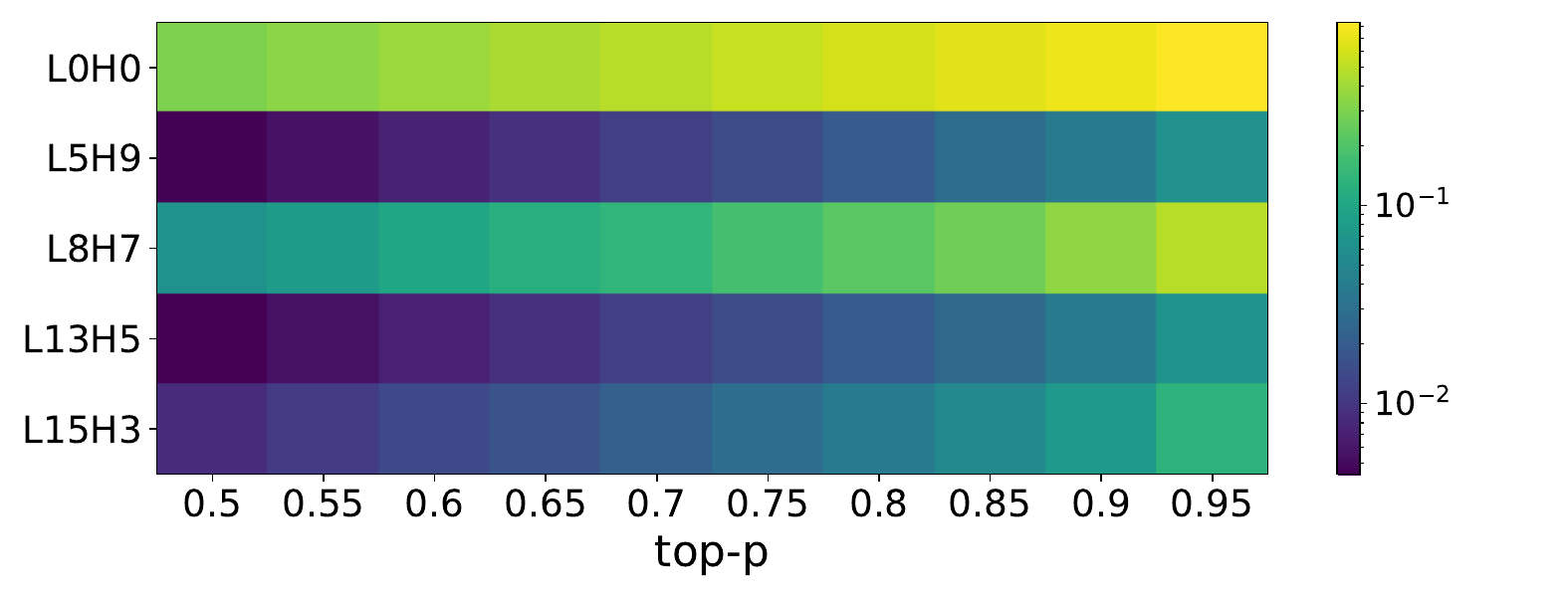}
  }
  \hfill
  \subfloat[
    \label{fig:iterations}
  ]{
    \includegraphics[width=0.62\linewidth]{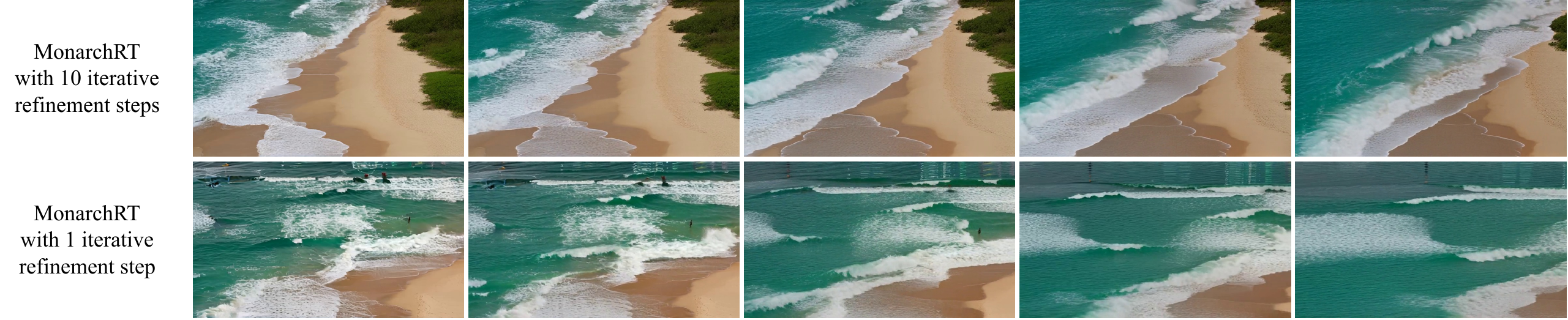}
  }
  \caption{\textbf{Left:} Percent of key-value tokens that fall under the top-$p$ threshold per head in Self-Forcing, averaged across all query tokens for several decoding iterations, on the first denoising iteration. Results shown for 5 randomly sampled heads/layers. textbf{Right:} Example generations on the same prompt on Self-Forcing using \Sys with 10 and 1 iterations of iterative refinement. 10 iterations has much higher quality but is practically inefficient, so we recover the accuracy of 1 iteration with training.}
  \label{fig:self-forcing}
\end{figure*}

\section{Observations and Analysis}

\label{sec:observation}
In this section, we present our core argument that, in principle, Monarch parameterization provides a fundamentally stronger approximation than sparse parameterization for modeling 3D attention. We further identify three key challenges that arise in practice, each of which limits its flexibility and achievable accuracy.

\subsection{Approximation Error Analysis}
\label{sec:topkfail}
\paragraph{Failure of oracle top-$k$ attention.} In~\Cref{fig:motivating_sparsity_error_comparison,fig:motivating_example_generations}, we show that oracle top-$k$ attention incurs large errors on approximating the attention map for high sparsity levels, leading to end-to-end quality degradation (the example car front exhibits severe geometric distortion), even with a $10\%$ computation budget (which is reasonably high for an oracle approximation). We attribute this to the 3D attention map being less sparse than expected. As illustrated in~\Cref{fig:attnscores}, for certain attention heads, more than $48\%\sim84\%$ tokens are required to recover $95\%$ attention score. For non-oracle sparse parameterizations, such as those based on position or block top-$k$, the situation will become even worse, as such methods are not guaranteed to retrieve the keys producing the highest attention scores.  
\begin{figure}[t]
  \centering
  \begin{subfigure}[t]{0.55\textwidth}
    \centering
    \includegraphics[width=\linewidth]{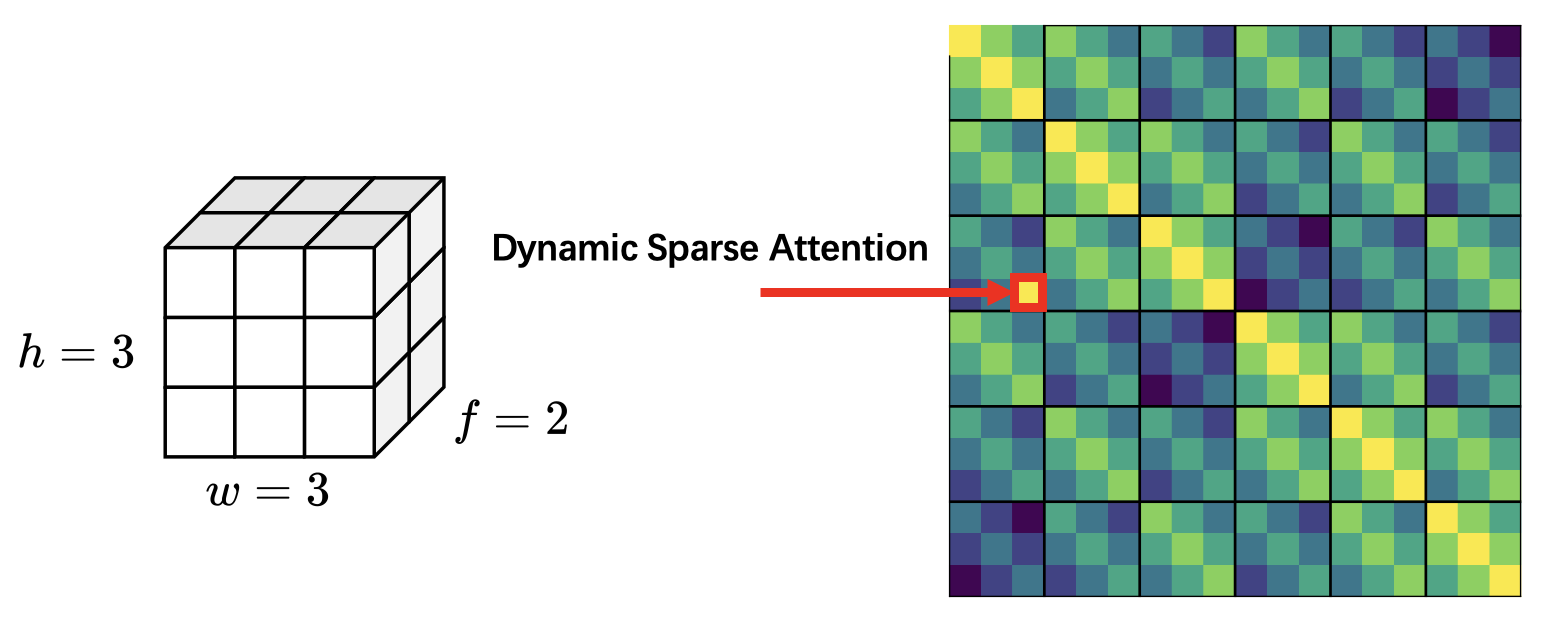}
  \end{subfigure} \hfill
  \caption{An illustration of our modeling of the 3D attention map in \Cref{eq:model}. \textbf{Left:} the shape of the video. \textbf{Right:} an example attention map. The periodic diagonal bands arise from spatiotemporal positional structure, 
while the large activation at position $(8,2)$ reflects a semantic relationship that is independent of position, requiring dynamic (retrieval-based) sparse attention to capture.}
  \label{fig:attnmodel}
\end{figure}

\subsection{Rethinking the Structure of 3D Attention Maps}
\label{sec:attentionmodel}
The failure of sparse parameterization stems from the incorrect assumption that the attention map is inherently sparse. 3D attention does not simply exhibit sparsity; instead, it reveals pronounced \emph{periodic structure} driven by spatiotemporal position, indicating that dense and repeating interactions are fundamental rather than exceptional. Inspired by~\citet{li2025radialattentiononlogn}, we informally model the attention map as
{
\begin{gather}
\label{eq:model}
\boldsymbol{A}_{(f_0,h_0,w_0),(f_1,h_1,w_1)} = \boldsymbol{D}_{(f_0,h_0,w_0),(f_1,h_1,w_1)} 
  + \boldsymbol{S}_{(f_0,h_0,w_0),(f_1,h_1,w_1)}
   + \epsilon, \\
\nonumber
\boldsymbol{D}_{(f_0,h_0,w_0),(f_1,h_1,w_1)}
= d_{w}(w_0,w_1)\;
   d_{h}(h_0,h_1)\;
   d_{t}(f_0,f_1).
\end{gather}
}

Here, $\boldsymbol{D}$ captures the \emph{positional} component of attention, modeled as the separable product of three distance functions along the spatial width (\(d_w\)), spatial height (\(d_h\)), and temporal (\(d_t\)) dimensions.  
Each distance function \(d(\cdot,\cdot) \le 1\) is monotonically decreasing, reflecting the empirical observation that attention scores induced by positional structure decay smoothly as tokens become farther apart.  
$\boldsymbol{S}$ represents the \emph{semantic} component, which models long-range relationships independent of positional proximity; most entries of $\boldsymbol{S}$ are $0$ due to the sparsity of meaningful semantic correspondence, the others are $1$.
\(\epsilon\) denotes residual noise or modeling error.  For simplicity, we omit the normalization here. This modeling immediately leads to~\Cref{thm:chunk_rank1}.

\begin{theorem}
\label{thm:chunk_rank1}
(informal) The 3D attention matrix \(\boldsymbol{A} \in \mathbb{R}^{fhw \times fhw}\) defined above admits a structural decomposition
\[
\boldsymbol{A} = \boldsymbol{P}\boldsymbol{D'} + \boldsymbol{S} + \epsilon
\]
where \(\boldsymbol{P}\) is a permutation matrix, \(\boldsymbol{D'}\) is \emph{blockwise rank-1} with block sizes $(b_1, b_2)$, which satisfies $b_1b_2 = fhw$ and \(\boldsymbol{S}\) is a sparse matrix.
\end{theorem}

\noindent Naively combining low-rank and sparse parameterization~\citep{zhang2025slasparsitydiffusiontransformers,chen2021scatterbrain,dong2024get} is not sufficient, because a matrix that is low-rank within each local block can still become full-rank when viewed globally.  We formally present \Cref{thm:chunk_rank1} in Appendix~\ref{sec:proofs}.

\subsection{Monarch Well-Represents 3D Attention Maps}
\label{sec:monarchfit}
\begin{figure}[t]
  \centering
  \begin{subfigure}[t]{0.7\textwidth}
    \centering
    \includegraphics[width=\linewidth]{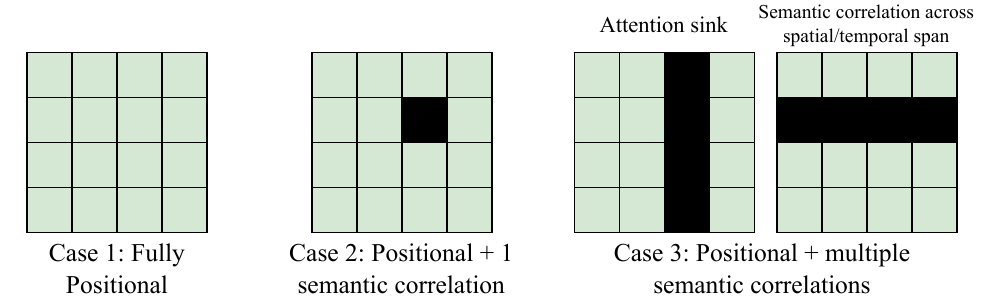}
  \end{subfigure} \hfill
  \caption{Illustration of several representative cases regarding occurrences of sparse, semantic correlations from a per-block ($\boldsymbol{\widetilde{A}_{[i,j]}}$) view of the full Monarch-parameterized attention map ($\boldsymbol{A}$).}
  \label{fig:rank1cases}
\end{figure}

In this section, we discuss the intuition that Monarch parameterization could effectively represent the 3D attention map, i.e., $\boldsymbol{A} = \boldsymbol{P}\boldsymbol{D'} + \boldsymbol{S} + \epsilon$.  
We consider each block
\[
\boldsymbol{\widetilde{A}_{[i,j]}} = \boldsymbol{{D'}_{[i,j]}} + \boldsymbol{\widetilde{S}_{[i,j]}} + \epsilon \in \mathbb{R}^{b_1 \times b_2},
\]
where $\boldsymbol{\widetilde{A}} = \boldsymbol{P}^\top \boldsymbol{A}$, $\boldsymbol{\widetilde{S}} = \boldsymbol{P}^\top \boldsymbol{S}$, $[i,j]$ indexes the block positions in the partitioned attention matrix (i.e. $\boldsymbol{\widetilde{A}_{[i,j]}} = \boldsymbol{\widetilde{A}}_{[ib_1:(i+1)b_1, jb_2:(j+1)b_2]}$), and $b_1, b_2$ denote the block sizes. We now intuitively analyze three representative cases, also illustrated in ~\Cref{fig:rank1cases}. 

\noindent \textbf{Case 1.} If $\mathrm{NNZ}(\boldsymbol{\widetilde{S}_{[i,j]}}) = 0$, the block is entirely governed by the positional term.  
As the positional structure factorizes along the three dimensions, $\boldsymbol{\widetilde{A}_{[i,j]}}$ reduces to a rank-1 matrix.

\noindent \textbf{Case 2.} When $\mathrm{NNZ}(\boldsymbol{\widetilde{S}_{[i,j]}}) = 1$ and the magnitude of $D_{[i,j]}$ is negligible, the block is dominated by a single semantic interaction between two distant tokens.  
In this situation, $\boldsymbol{\widetilde{A}_{[i,j]}}$ is effectively determined by the single nonzero entry in $\boldsymbol{\widetilde{S}_{[i,j]}}$, and the remaining entries can be approximated as zero.  
Consequently, $\boldsymbol{\widetilde{A}_{[i,j]}}$ is also rank-1.

\noindent \textbf{Case 3.} When $\mathrm{NNZ}(\boldsymbol{\widetilde{S}_{[i,j]}}) > 1$ but all nonzero entries lie within a single row or a single column, while others are negligible the block corresponds to multiple semantic connections originating from (or pointing to) the same token. This pattern includes the common attention-sink phenomenon.  Since all semantic interactions are confined to one row or column, $\boldsymbol{\widetilde{A}_{[i,j]}}$ remains rank-1.

Since the number of strong semantic interactions (the nonzero entries of \(\boldsymbol{S}\)) is limited, an appropriate permutation $\boldsymbol{P}$ and block sizes $(b_1, b_2)$ can ideally prevent multiple semantic entries from falling into the same block or from mixing with strong positional interactions (i.e., from falling out of the listed 3 cases).  Therefore, the 3D attention map $\boldsymbol{A}$ can be represented by Monarch matrices with permutation. We present a more detailed analysis of Monarch parameterizations on attention in Appendix~\ref{sec:expressivenessmonarchattention}.  

\begin{figure}[t]
  \centering
  \begin{subfigure}[t]{0.7\textwidth}
    \centering
    \includegraphics[width=\linewidth]{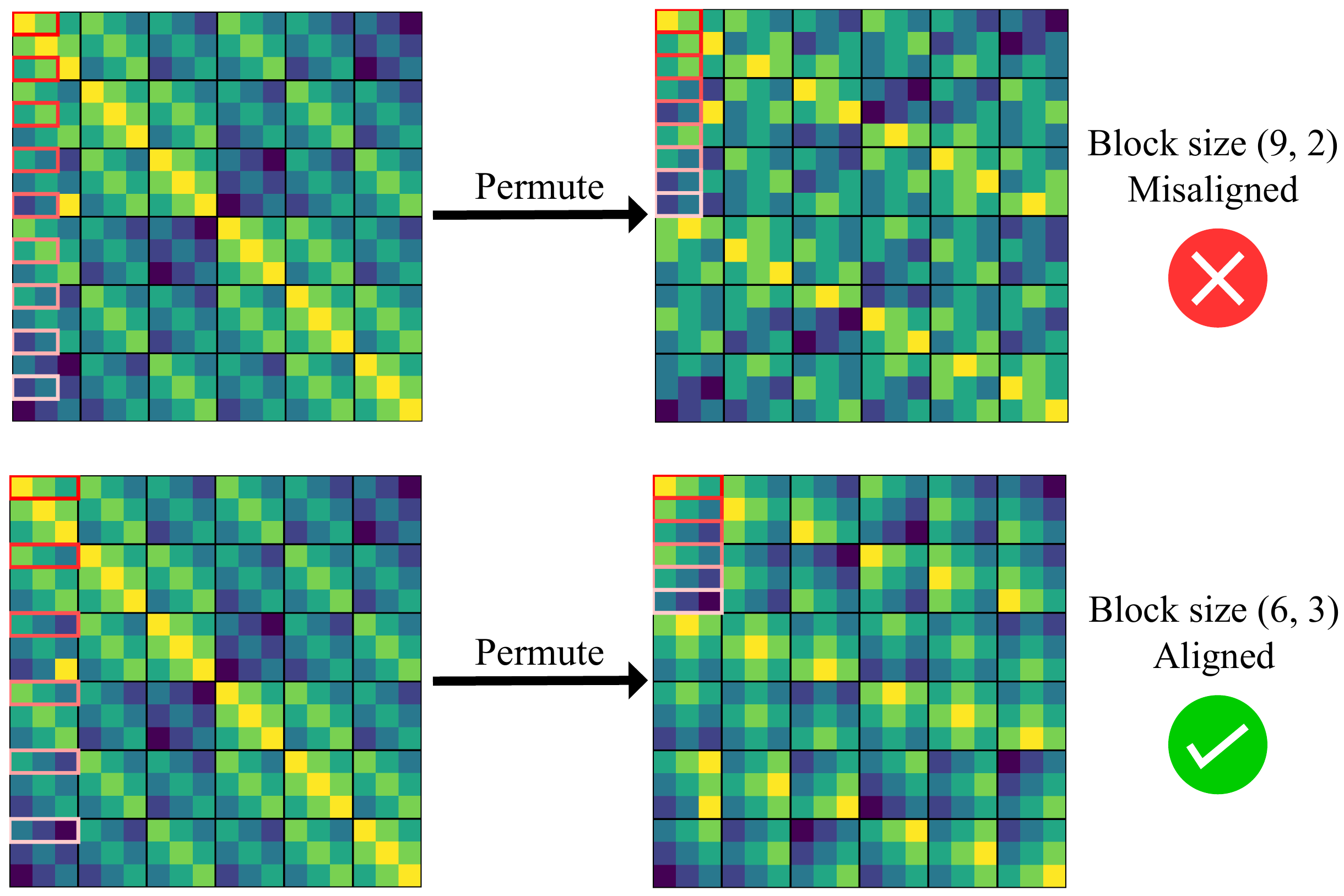}
  \end{subfigure} \hfill
  \caption{The comparison of two Monarch factorizations. The attention map is the same as~\Cref{fig:attnmodel}. \textbf{Top:} By selecting misaligned block sizes, most of the blocks are obviously not rank-1, losing the ability to represent positional relationship. \textbf{Bottom:} With aligned block sizes, the factorization is able to recover positional relationship.}
  \label{fig:blkanalysis}
\end{figure}

\subsection{Practical Challenges of MonarchAttention in Video Generation}
\label{sec:challenges}
Aside from the advantages, we identify three challenges that prevent direct usage of MonarchAttention in video generation models.

\subsubsection{Challenge 1: Block misalignment with spatiotemporal structure.}
The effectiveness of Monarch parameterization critically depends on whether its block structure
aligns with the underlying spatiotemporal organization of video tokens.
We illustrate this in~\Cref{fig:blkanalysis} using a $3$D attention example with $(f,h,w)=(2,3,3)$, yielding an
$18 \times 18$ attention matrix, and compare two block configurations.

In the first configuration, we choose block sizes $(b_1,b_2)=(6,3)$, which group tokens that are
spatially adjacent within the same frame, such as
$x_{t,i,0}$, $x_{t,i,1}$, and $x_{t,i,2}$.
This grouping respects the natural spatiotemporal locality of the video, and as a result,
nearly all blocks exhibit an approximately rank-$1$ structure, leading to a high-quality
approximation.

In contrast, choosing block sizes $(9,2)$ produces blocks of the same total size but groups tokens
according to their flattened indices.
This grouping mixes tokens that are distant in the original spatiotemporal layout, even though
they appear close after flattening.
Consequently, most blocks no longer admit a low-rank structure, and the approximation quality
degrades sharply.
As shown in~\Cref{fig:motivating_example_generations}, aligned block grouping preserves visual fidelity,
whereas misaligned grouping leads to severe quality degradation.

This example highlights that, unlike 1D sequences, flattened token order in 3D video does not
reflect true spatial or temporal proximity, and that improper block alignment can fundamentally
limit the effectiveness of Monarch parameterization.

\subsubsection{Challenge 2: Lack of monotonic refinement with increased computation.}
Even when block sizes are perfectly aligned with the spatiotemporal structure, semantic
interactions may still cause certain blocks to deviate from being rank-$1$.
Crucially, such semantic correspondences are sparse and irregular, making their locations
inherently unpredictable.
A natural strategy to improve approximation accuracy is therefore to increase computational
budget, e.g., by using smaller blocks that are less likely to mix multiple semantic interactions.

However, in the original Monarch parameterization, increasing computation does not reliably
translate into better approximations.
Specifically, Monarch enforces the constraint $b_1 b_2 = N$, where $N=fhw$ is the total number of
tokens.
Under this constraint, the total number of parameters in the Monarch factors scales as
$\mathcal{O}(N(b_1 + b_2))$.
As a result, changing block sizes $(b_1,b_2)$ redistributes parameters between the two factors
$\boldsymbol{L}$ and $\boldsymbol{R}$, but does not guarantee an increase in total computation.

Consequently, the constraint $b_1 b_2 = N$ prevents a monotonic refinement: increasing compute by modifying $(b_1,b_2)$ necessarily trades finer partitioning in one
dimension for coarser partitioning in the other, so semantic interactions remain mixed within
blocks and the approximation error may not improve.

This contrasts sharply with sparse-attention-based methods (like topk-$k$), where additional computation directly
corresponds to attending to more tokens and thus yields a monotonic accuracy--efficiency trade-off.

\subsubsection{Challenge 3: High overhead of iterative refinement.}
MonarchAttention estimates the Monarch factors $\boldsymbol{L}$ and $\boldsymbol{R}$ through an
iterative refinement procedure.
While increasing the number of refinement steps generally improves approximation accuracy, it
introduces substantial additional computation.
Specifically, the runtime of MonarchAttention scales linearly with the number of
iterations, directly reducing throughput.

%% file: sections/method.tex
\section{MonarchRT}
\label{sec:method}
In this section, we propose \Sys by addressing the three challenges identified above, i.e., how to align block sizes effectively (\Cref{sec:grouping}), how to enable finer-grained Monarch parameterization beyond fixed block areas (\Cref{sec:tiled}), and how to reduce the parameterization cost for real-time usage (\Cref{sec:implementation}). We then describe our custom implementation that supports efficient training of MonarchAttention on long sequences.

\subsection{Aligning Monarch Blocks}
\label{sec:grouping}

\textbf{Key insight:} A Monarch parameterization is \emph{aligned} with video attention if each
spatiotemporal dimension ($f$, $h$, $w$) is entirely contained within exactly one block dimension.
Only under this condition can Monarch exactly represent fully separable positional attention
patterns.

To formalize this notion, we consider a simplified setting where the attention follows the purely
positional model in~\Cref{eq:model} by temporarily assuming $\boldsymbol{S}=0$.
Recall that a Monarch matrix is parameterized by two block-diagonal factors
$\boldsymbol{L}$ and $\boldsymbol{R}$, which can be interpreted as 3D tensors of shape
$b_2 \times b_1 \times b_1$ and $b_1 \times b_2 \times b_2$, respectively, with block sizes
$b_1$ and $b_2$ satisfying $b_1 b_2 = fhw$.

\paragraph{An aligned construction.}
Consider choosing block sizes $b_1 = fh$ and $b_2 = w$.
In this case, $\boldsymbol{L}$ has shape $w \times fh \times fh$ and $\boldsymbol{R}$ has shape
$fh \times w \times w$.
Under this parameterization, the post-softmax attention scores can be written as
\[
\boldsymbol{A}_{(f_0,h_0,w_0),(f_1,h_1,w_1)}
= \boldsymbol{L}_{w_0,(f_0,h_0),(f_1,h_1)}
  \boldsymbol{R}_{(f_1,h_1),w_0,w_1}.
\]
By setting
\begin{gather*}
\boldsymbol{L}_{w_0,(f_0,h_0),(f_1,h_1)} = d_t(f_0,f_1)\, d_h(h_0,h_1),\\
\boldsymbol{R}_{(f_1,h_1),w_0,w_1} = d_w(w_0,w_1),
\end{gather*}
the Monarch parameterization exactly reproduces the fully separable positional attention model in
\Cref{eq:model}.
Importantly, any permutation of the dimensions across the block sizes, such as
$(b_1,b_2)=(f,hw)$ or $(b_1,b_2)=(fw,h)$, admits an equivalent exact decomposition.

\paragraph{When alignment fails.}
Crucially, no single dimension ($f$, $h$, or $w$) may be split across both block sizes.
If a dimension partially spans $b_1$ and $b_2$, it becomes impossible to factor the attention into
fully separable terms $\boldsymbol{L}$ and $\boldsymbol{R}$.
As a result, the Monarch approximation cannot exactly represent the assumed attention structure,
even with increased parameter count.
This explains the behavior observed in~\Cref{fig:motivating_example_generations}, where aligned block sizes
(e.g., $(fh,w)$) preserve visual fidelity, while misaligned choices
(e.g., $(f \cdot \tfrac{h}{4},\,4w)$) lead to severe degradation despite higher nominal capacity.

\begin{tcolorbox}[colback=white,colframe=softlavender,title=Take Away]
We therefore define a Monarch parameterization for video attention to be \emph{aligned} if each of
the video dimensions $f$, $h$, and $w$ is fully assigned to exactly one block dimension.
Excluding the degenerate dense cases $(fhw,1)$ and $(1,fhw)$, this yields exactly six aligned block
configurations:
\[
(fh,w),\ (w,fh),\ (f,hw),\ (hw,f),\ (fw,h),\ (h,fw).
\]
\end{tcolorbox}

\subsection{Tiled Monarch Parameterization}
\label{sec:tiled}

We begin by recalling the standard Monarch parameterization.
Given block sizes $(b_1,b_2)$ with $N=b_1b_2$, a Monarch matrix $\boldsymbol{M}\in\mathbb{R}^{N\times N}$
can be written as
\[
\boldsymbol{M}_{mn}
= \boldsymbol{M}_{(\ell b_2 + j)(k b_2 + i)}
= \boldsymbol{L}_{j\ell k}\,\boldsymbol{R}_{kji},
\]
where $\boldsymbol{L}\in\mathbb{R}^{b_2\times b_1\times b_1}$ and
$\boldsymbol{R}\in\mathbb{R}^{b_1\times b_2\times b_2}$.

\paragraph{Algorithm.}
We propose \emph{Tiled Monarch Parameterization}, which generalizes Monarch by decomposing each
Monarch block into smaller sub-blocks.
Specifically, we introduce integers $c_1$ and $c_2$ such that $c_1 \mid b_1$ and $c_2 \mid b_2$.
Instead of a single Monarch factorization, we represent $\boldsymbol{M}$ as a collection of
$c_1^2 c_2^2$ Monarch tiles, each with block sizes
$(\tfrac{b_1}{c_1}, \tfrac{b_2}{c_2}$).

Formally, we parameterize $\boldsymbol{M}$ using tiled factors
$\boldsymbol{L}'$ and $\boldsymbol{R}'$, where
$\boldsymbol{L}'$ has shape
$(c_1, c_2, c_1, c_2, \tfrac{b_2}{c_2}, \tfrac{b_1}{c_1}, \tfrac{b_1}{c_1})$
and $\boldsymbol{R}'$ has shape
$(c_1, c_2, c_1, c_2, \tfrac{b_1}{c_1}, \tfrac{b_2}{c_2}, \tfrac{b_2}{c_2})$.
Each tile independently parameterizes a local rank-$1$ structure.

To estimate tiled Monarch factors for the attention matrix $\boldsymbol{A}$, we extend
MonarchAttention by applying the same alternating refinement procedure independently to each tile.
To enforce row-stochasticity, we impose slightly stronger constraints on the tiled factors, detailed
in~\Cref{sec:modifiedmonarchattn}.

\begin{theorem}[Strict expressiveness of tiled Monarch]
\label{thm:tiled_monarch_strict}
Fix base block sizes \((b_1,b_2)\) with \(N=b_1b_2\), and let \(c_1\mid b_1\), \(c_2\mid b_2\) be tiling factors.
Let \(\mathcal{M}(b_1,b_2)\) denote the set of Monarch matrices with block sizes \((b_1,b_2)\), i.e.,
matrices \(\boldsymbol{M}\in\mathbb{R}^{N\times N}\) that admit factors
\(\boldsymbol{L}\in\mathbb{R}^{b_2\times b_1\times b_1}\),
\(\boldsymbol{R}\in\mathbb{R}^{b_1\times b_2\times b_2}\) satisfying
\[
\boldsymbol{M}_{(\ell b_2 + j)(k b_2 + i)}=\boldsymbol{L}_{j\ell k}\,\boldsymbol{R}_{kji}.
\]
Let \(\mathcal{M}_{\mathrm{tile}}(b_1,b_2;c_1,c_2)\) denote the set of matrices representable by the tiled Monarch
parameterization with the same base block sizes \((b_1,b_2)\) and tiling factors \((c_1,c_2)\).
Then
\[
\mathcal{M}(b_1,b_2)\subseteq \mathcal{M}_{\mathrm{tile}}(b_1,b_2;c_1,c_2).
\]
Moreover, if \(c_1>1\) or \(c_2>1\), the inclusion is strict:
\[
\mathcal{M}(b_1,b_2)\subset \mathcal{M}_{\mathrm{tile}}(b_1,b_2;c_1,c_2),
\]
i.e., every (untiled) Monarch matrix can be represented exactly by a tiled Monarch matrix with appropriate parameter
tying, but there exist tiled Monarch matrices that cannot be represented by any (untiled) Monarch parameterization
with block sizes \((b_1,b_2)\).
\end{theorem}

In other words, tiled Monarch is strictly more expressive than regular Monarch for the same base block sizes. We formally prove~\Cref{thm:tiled_monarch_strict} in Appendix~\ref{sec:tiled_monarch_proof}. Informally, though, tiled factors increase parameter counts by factors of $c_2$ and $c_1$ respectively, for $\boldsymbol{L}$ and $\boldsymbol{R}$. By tying parameters across tiles, one can exactly recover the original Monarch parameterization, so tiled Monarch is intuitively at least as expressive as standard Monarch.

Crucially, tiled Monarch enables \emph{controllable refinement}.
By increasing $(c_1,c_2)$, each original block is subdivided into smaller sub-blocks, allowing the
approximation to better capture sparse and irregular semantic interactions that cannot be modeled
by a single rank-$1$ block.

\paragraph{Example.}
Recall that for video attention with resolution $(f,h,w)$, we set the aligned Monarch block sizes
to $(b_1,b_2)=(fh,w)$ as shown in~\Cref{sec:grouping}.
However, when attention exhibits neighborhood-level sparsity with neighborhood size
$(n_f,n_h,n_w)$, a single Monarch block may span multiple neighborhoods and therefore violate the
rank-$1$ assumption.

Tiled Monarch resolves this issue by enforcing locality within each tile.
Specifically, we choose
\[
c_1 = \frac{f}{n_f}\cdot\frac{h}{n_h},
\qquad
c_2 = \frac{w}{n_w},
\]
resulting in $c_1 c_2$ Monarch tiles, each with block sizes $(n_f n_h, n_w)$.
Each tile therefore contains tokens from only a single spatiotemporal neighborhood along every
dimension, making the rank-$1$ assumption locally valid.

The degenerate case $(n_f,n_h,n_w)=(1,1,1)$ reduces to dense attention.
Empirically, we find that choosing $n_f$ as a small constant while letting
$n_h=\mathcal{O}(h)$ and $n_w=\mathcal{O}(w)$ achieves high visual quality with a sparse
parameterization.

\paragraph{Computational complexity.}
Tiled Monarch preserves the efficient computation pattern of MonarchAttention.
Each tile is processed independently using the same alternating refinement procedure, and the
overall complexity scales linearly with the number of tiles.

\begin{tcolorbox}[colback=white,colframe=softlavender,title=Take Away]
In standard Monarch parameterization, changing block shapes does not reliably control either computation cost or expressiveness.
In contrast, Tiled Monarch parameterization enables fine-grained block structures and, similar to top-$k$ sparse attention methods,
provides a clear and monotonic trade-off between accuracy and efficiency.
\end{tcolorbox}

\subsection{Finetuning and Efficient Implementation}
\label{sec:implementation}

\begin{figure}[t]
  \centering
  \begin{subfigure}[t]{\textwidth}
    \centering
    \includegraphics[width=\linewidth]{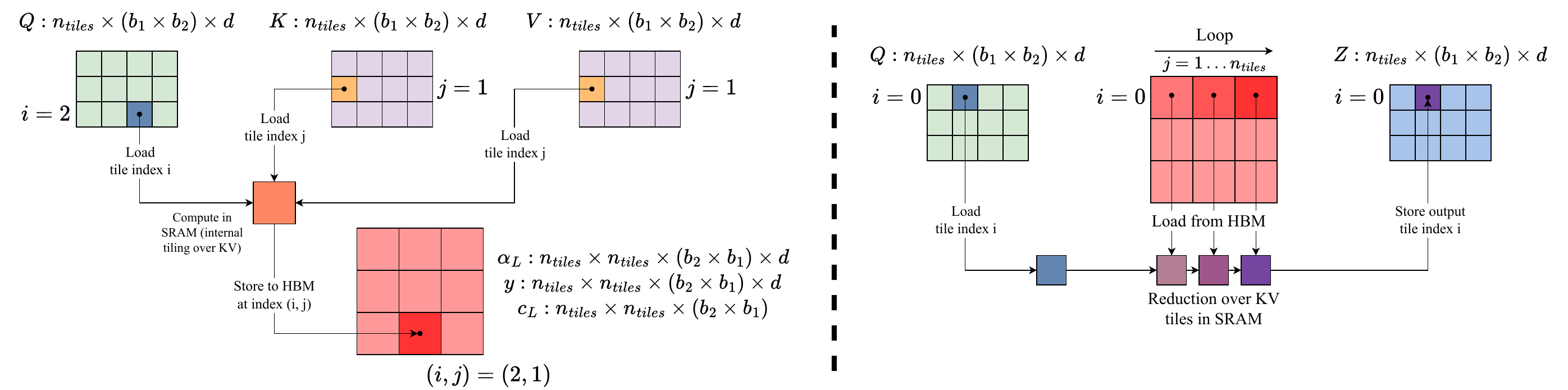}
  \end{subfigure} \hfill
  \caption{Visualization of efficient kernel implementation for MonarchAttention algorithm with a tiled Monarch parameterization. \textbf{Left}: first stage to compute $\alpha_L, y, c_L$ tiles. \textbf{Right}: second stage (separate kernel) to compute output by reducing over KV tiles.}
  \label{fig:kernel_vis}
\end{figure}

To further reduce the computational overhead of iterative refinement, we introduce \emph{Monarch finetuning}, a lightweight training procedure that dramatically decreases the number of refinement steps required to obtain high-quality Monarch parameterizations. In practice, we find that with finetuning, even a single refinement iteration is often sufficient to match the visual fidelity of much more expensive multi-step optimization.

To enable end-to-end training of Monarch factors, we implement custom forward and backward kernels tailored for long-sequence 3D attention. We visualize the forward process in~\Cref{fig:kernel_vis} and provide additional information on the MonarchAttention algorithm itself in~\Cref{sec:origmonarchattn,sec:modifiedmonarchattn}. Similar to the original MonarchAttention, the $\boldsymbol{\beta}$ terms---and hence the resulting $\boldsymbol{L}$ and $\boldsymbol{R}$ factors---can be materialized directly in SRAM using a FlashAttention-style computation pattern. However, the $\boldsymbol{\alpha}$ and $\mathbf{c}$ terms must be materialized in HBM, as their tensor shapes depend jointly on $f_q$ and $f_{kv}$. For full non-causal 3D attention, this becomes quadratic in the number of frames and would ordinarily be prohibitive.

To overcome this limitation, we adopt a \emph{mini-sequence} strategy: the query frames are divided into smaller chunks, and the complete attention output for each chunk is computed before moving on to the next. This is valid because query frames are independent during the attention computation, allowing us to cap peak memory usage without altering correctness. Since 3D attention workloads are strongly compute-bound, this chunked processing introduces negligible overhead while enabling scalable fintuning of Monarch parameterizations.

%% file: sections/experiment.tex
\section{Empirical Validation}

\label{sec:experiments}
In this section, we demonstrate that \Sys can speed up on top of the SOTA real-time video generation models while preserving video fidelity. We first present \Sys's generation quality on downstream tasks, then an end-to-end evaluation of \Sys's throughput. 
\begin{itemize}
    \item In~\Cref{sec:quality_eval}, we demonstrate \Sys preserves generation quality for real-time video generations for Self-Forcing~\citep{huang2025selfforcingbridgingtraintest} with computation cost as little as $\sim5\%$ of full attention, surpassing existing sparse attention based algorithms.
    \item in~\Cref{sec:training_free_ablation}, we conduct further training-free ablations and find again that, even with $\sim5\%$ attention density, \Sys continues to surpass additional sparse attention baselines, including oracle top-$k$ attention.
    \item In~\Cref{sec:efficiency_evaluations}, we conduct efficiency evaluations and find that our efficient kernel can provide speedups in the range of $1.4-11.8\times$ speedups to attention compared to various FlashAttention kernels ~\citep{shah2024flashattention3fastaccurateattention}. On RTX 5090, we achieve (for the first time) true 16 FPS real-time generation with high quality on Self-Forcing.
\end{itemize}

\subsection{Quality Evaluations}
\label{sec:quality_eval}

\begin{table*}
\centering
\renewcommand{\arraystretch}{1.2}
\begin{adjustbox}{max width=\width}
\begin{tabular}{@{} c c c c @{}}
\toprule
Method
& Quality Score
& Semantic Score
& Total Score \\
\midrule
Dense Attention & 0.844 & 0.804 & 0.836 \\
\Sys (95\% sparse) & 0.846 & 0.805 & 0.838 \\
\bottomrule
\end{tabular}
\end{adjustbox}

\caption{VBench scores for base model and trained \Sys on Self-Forcing.}
\label{tab:trained_self_forcing}
\end{table*}

\begin{table*}
\centering
\scriptsize
\renewcommand{\arraystretch}{1.2}
\begin{adjustbox}{max width=\width}
\begin{tabular}{@{} c c c c c c c @{}}
\toprule
& \multicolumn{3}{c}{\textbf{4-step}} & \multicolumn{3}{c}{\textbf{50-step}} \\
\cmidrule(lr){2-4}\cmidrule(lr){5-7}
Method
& Quality Score
& Semantic Score
& Total Score
& Quality Score
& Semantic Score
& Total Score \\
\midrule
Dense Attention & \textbf{0.846} & \textbf{0.800} & \textbf{0.837} & \textbf{0.846} & 0.810 &\textbf{ 0.839} \\
VSA (85\% sparse) & 0.828 & 0.793 & 0.821 & 0.827 & 0.785 & 0.819 \\
\Sys (95\% sparse) & 0.842 & 0.788 & 0.832 & 0.841 & \textbf{0.812} & 0.835 \\
\bottomrule
\end{tabular}
\end{adjustbox}
\caption{Quality evaluation for a Wan2.1-1.3B (base 50-step and distilled 4-step models) for dense attention as well as trained VSA and \Sys. Higher values indicate better quality for all metrics.}
\label{tab:trained_wan}
\end{table*}

We demonstrate that \Sys can preserve the generation quality in diverse tasks with as low as $\sim5\%$ attention density.

\noindent\textbf{Setup.} We conduct training-free evaluations with VBench~\citep{huang2023vbenchcomprehensivebenchmarksuite}. In order to emphasize the effectiveness of \Sys for auto-regressive and few-step diffusion models, we evaluate \Sys on Self-Forcing~\citep{huang2025selfforcingbridgingtraintest}, which is finetuned from Wan 2.1-1.3B~\citep{wan2025wanopenadvancedlargescale} while applying auto-regressive generation and DMD~\citep{yin2024onestepdiffusiondistributionmatching,yin2024improveddistributionmatchingdistillation} to accommodate real-time generation. We also show results for both Wan 2.1-1.3B (a 50-step bidirectional model) as well as a version of this model that we distilled using DMD to 4 steps.

For Self-Forcing, we inject \Sys directly into the DMD stage of the Self-Forcing training pipeline rather than finetuning on top of the dense Self-Forcing checkpoint. For the 4-step distilled Wan model, we similarly inject \Sys directly into the DMD stage rather than finetuning directly on top of the distilled dense model. For 50-step Wan, we directly apply diffusion loss finetuning on the base model. 

\noindent\textbf{Baselines.} We mainly evaluate our training-based method against dense baselines as well as VSA~\citep{zhang2025vsafastervideodiffusion}, another training-based dynamic sparse attention method, although we do not show VSA results for Self-Forcing as it only supports full bidirectional attention. We compare with additional baselines under a training-free setting in~\Cref{sec:training_free_ablation}.

\noindent\textbf{Main results and analysis.} The results in~\Cref{tab:trained_self_forcing} show that \Sys remains close in performance to the dense model for both 1.3B and 14B models, even at 95\% effective sparsity. Table~\Cref{tab:trained_wan}, we observe similar results for bidirectional Wan (both 4-step and 50-step) and find that \Sys outperforms VSA on all metrics, even though it is evaluated at a higher sparsity level.

\subsection{Training-Free Ablations}
\label{sec:training_free_ablation}

\begin{table*}
\centering
\begin{tabular}{@{}cccc@{}}
\toprule
Method & \shortstack{Quality\\Score} & \shortstack{Semantic\\Score} & \shortstack{Total\\Score} \\
\midrule
\shortstack{Dense Attention} & 0.844 & 0.804 & 0.836 \\
\shortstack{Exact top-$k$ (85\% sparse)} & 0.834 & 0.658 & 0.799 \\
\shortstack{SVG (85\% sparse)} & 0.715 & 0.214 & 0.615 \\
\shortstack{RadialAttention (85\% sparse)} & 0.841 & 0.718 & 0.816 \\
\shortstack{\Sys (90\% sparse)} & \textbf{0.847} & \textbf{0.808} & \textbf{0.839} \\
\bottomrule
\end{tabular}%
\caption{VBench evaluation for Self-Forcing, with all sparse methods evaluated training-free. While SVG and RadialAttention are reported as using an 85\% sparse mask, this sparsity level is an overestimate, as the SVG/RadialAttention masks are combined with the autoregressive block-causal mask (and the block-causal mask does not count towards the sparsity level). For RadialAttention and SVG, we also retain the first denoising step (out of 4 for Self-Forcing) and first attention
block as dense attention.}
\label{tab:training_free_self_forcing}
\end{table*}

\begin{table*}
\centering
\scriptsize
\renewcommand{\arraystretch}{1.2}
\begin{adjustbox}{max width=\width}
\begin{tabular}{@{} c c c c c c c c @{}}
\toprule
& \multicolumn{4}{c}{\textbf{4-step}} & \multicolumn{3}{c}{\textbf{50-step}} \\
\cmidrule(lr){2-5}\cmidrule(lr){6-8}
Method
& PSNR ($\uparrow$)
& SSIM ($\uparrow$)
& LPIPS ($\downarrow$)
& VBench ($\uparrow$)
& PSNR ($\uparrow$)
& SSIM ($\uparrow$)
& LPIPS ($\downarrow$) \\
\midrule
Dense Attention
& -- & -- & -- & \textbf{0.846}
& -- & -- & -- \\
SVG (85\% sparse)
& 9.411 & 0.203 & 0.749 & 0.746
& 11.349 & 0.286 & 0.678 \\
SVG2 ($\sim$85\% sparse)
& 11.154 & 0.321 & 0.631 & 0.823
& 14.625 & 0.438 & 0.537 \\
SVG2 ($\sim$90\% sparse)
& 10.737 & 0.307 & 0.662 & 0.808
& 14.116 & 0.417 & 0.568 \\
RadialAttention (85\% sparse)
& 11.427 & 0.290 & 0.711 & 0.727
& 13.719 & 0.329 & 0.674 \\
\Sys\ (90\% sparse)
& \textbf{12.657} & \textbf{0.364} & \textbf{0.585} & 0.834
& \textbf{17.220} & \textbf{0.525} & \textbf{0.506} \\
\bottomrule
\end{tabular}
\end{adjustbox}
\caption{Quality evaluation for a Wan2.1-1.3B (base 50-step and distilled 4-step models). All sparse methods are evaluated training-frees. Sparsity levels for SVG2 are average estimates, as the exact sparsity level is not explicitly controllable in SVG2.}
\label{tab:training_free_wan}
\end{table*}

To further demonstrate the effectiveness of Monarch parameterization, we evaluate \Sys against several additional baselines all in a training-free setting.

\textbf{Setup.} We conduct training-free evaluations with VBench~\citep{huang2023vbenchcomprehensivebenchmarksuite} primarily on Self-Forcing and distilled Wan 2.1-1.3B (4-step).

\textbf{Baselines.} We evaluate the quality of our method against several baselines, including full dense attention and oracle top-$k$ attention. We also include existing SOTA sparse attention methods, namely Sparse VideoGen~\citep{xi2025sparsevideogenacceleratingvideo}, Sparse VideoGen-2~\citep{yang2025sparse}, and RadialAttention~\citep{li2025radialattentiononlogn}. As these sparse attention methods are only proposed for full bidirectional attention, we mainly compare them with \Sys for Wan. However, we additionally adapt Sparse VideoGen (SVG) and RadialAttention to Self-Forcing by combining the autoregressive causal mask with their respective static attention masks.

\textbf{Main results and analysis.} As shown in Table \ref{tab:training_free_self_forcing}, SVG fails to produce coherent output by most metrics --- notably, the result is produced using SVG with ``warmup", i.e. maintaining dense attention for the first timestep (out of a total 4 steps for Self-Forcing) as well as keeping the first layer as dense attention for all timesteps. The results also show that RadialAttention and oracle top-$k$ both demonstrate notable quality reduction as well relative to the dense baseline, supporting our claim that sparsity is an inherently flawed approximation for video attention. However, training-free \Sys remains quite close to the dense model even with up to 90\% sparsity. Similarly,~\Cref{tab:training_free_wan} shows that \Sys achieves the highest performance over all other sparse methods for the 4-step distilled bidirectional Wan model, both on VBench as well as on other holistic metrics, namely PSNR, SSIM, and LPIPS~\citep{zhang2018unreasonableeffectivenessdeepfeatures} (we also include these additional metrics on the base 50-step Wan model for reference).

\subsection{Efficiency Evaluations}
\label{sec:efficiency_evaluations}

We demonstrate significant speedups for both Self-Forcing and Wan using our efficient Triton kernel implementation. In~\Cref{tab:kernel_wan,tab:kernel_sf,tab:e2e_wan,tab:e2e_sf}, we benchmark individual attention kernel latency as well as E2E latency on Nvidia RTX 5090 and H100 GPUs. We compare FA-2 / FA-3 / FA-4 (for RTX 5090 / H100 / B200) to VSA and \Sys at 480p and (theoretical) 720p resolution. We always evaluate \Sys at two sparsity levels, corresponding to block sizes $(h, w)$ and $(3h, w)$ respectively (the latter resulting from $n_f = 3$), where $h$ and $w$ depend on the latent dimensions for the given resolution. For VSA, we measure at $85\%$ sparsity as well as the two effective sparsity levels used for \Sys. Since VSA also requires pre- and post-processing steps, we use \texttt{torch.compile} on VSA to provide a fair evaluation.

\begin{table}[H]
\centering
\setlength{\tabcolsep}{3.2pt}
\begin{subtable}{\textwidth}
\centering
\begin{tabularx}{\textwidth}{c*{6}{Y}}
\toprule
\multirow{2}{*}{GPU}
& \multirow{2}{*}{\shortstack{Flash\\Attention}}
& \multicolumn{3}{c}{VSA} & \multicolumn{2}{c}{\Sys} \\
\cmidrule(lr){3-5}\cmidrule(lr){6-7}
&  & $s=0.85$ & $s=0.95$ & $s=0.97$
& $s=0.95$ & $s=0.97$ \\
\midrule
RTX 5090 & 31.15 & 13.52 &  8.55 &  7.89 &  6.76 &  3.39 \\
H100     &  9.74 &  6.48 &  4.95 &  4.60 &  6.24 &  2.61 \\
B200     &  4.53 &  5.69 &  4.02 &  3.65 &  5.97 &  2.41 \\
\bottomrule
\end{tabularx}
\caption{480p}
\end{subtable}

\vspace{4pt}

\begin{subtable}{\textwidth}
\centering
\begin{tabularx}{\textwidth}{c*{6}{Y}}
\toprule
\multirow{2}{*}{GPU}
& \multirow{2}{*}{\shortstack{Flash\\Attention}}
& \multicolumn{3}{c}{VSA} & \multicolumn{2}{c}{\Sys} \\
\cmidrule(lr){3-5}\cmidrule(lr){6-7}
&  & $s=0.85$ & $s=0.97$ & $s=0.98$
& $s=0.97$ & $s=0.98$ \\
\midrule
RTX 5090 & 159.11 & 55.51 & 24.03 & 19.84 & 28.24 & 13.53 \\
H100     &  53.29 & 23.93 & 13.46 & 12.15 & 18.61 &  9.59 \\
B200     &  24.78 & 21.71 & 11.26 & 9.86 & 18.04 & 9.56 \\
\bottomrule
\end{tabularx}
\caption{720p}
\end{subtable}

\caption{Attention kernel latency (ms) for workload of generating an 81-frame video with Wan 2.1-1.3B. Sparsity level $s$ is indicated for VSA and \Sys.}
\label{tab:kernel_wan}

\end{table}

\begin{table}[H]
\centering
\setlength{\tabcolsep}{4.2pt}

\begin{tabularx}{\columnwidth}{c*{6}{Y}}
\toprule
& \multicolumn{3}{c}{480p} & \multicolumn{3}{c}{720p} \\
\cmidrule(lr){2-4}\cmidrule(lr){5-7}
\multirow{2}{*}{GPU}
& \multirow{2}{*}{\shortstack{Flash\\Attention}}
& \multicolumn{2}{c}{\Sys}
& \multirow{2}{*}{\shortstack{Flash\\Attention}}
& \multicolumn{2}{c}{\Sys} \\
\cmidrule(lr){3-4}\cmidrule(lr){6-7}
&  & $s=0.95$ & $s=0.97$ &  & $s=0.97$ & $s=0.98$ \\
\midrule
RTX 5090 &  4.97 & 1.27 & 0.63 & 22.69 & 4.22 & 2.01 \\
H100     &  1.58 & 1.04 & 0.57 &  7.46 & 2.83 & 1.53 \\
B200     &  0.79 & 0.95 & 0.45 &  3.75 & 2.69 & 1.49 \\
\bottomrule
\end{tabularx}

\caption{Attention kernel latency (ms) for workload of decoding the final frame in an 81-frame video with Self-Forcing. Sparsity level $s$ is indicated for \Sys.}
\label{tab:kernel_sf}

\end{table}

In~\Cref{tab:kernel_wan,tab:kernel_sf}, we observe that \Sys achieves up to 9.2$\times$ attention speedup over FA-2 on RTX 5090 and up to 3.7$\times$ speedup over FA-3 on H100 at 480p resolution. At 720p resolution, these peak speedups become 11.8$\times$ and 5.6$\times$ respectively. \Sys is also able to achieve theoretical speedups over FA-4, including $\sim 1.4\times$ at 720p for Self-Forcing.

\begin{table}[H]
\centering
\setlength{\tabcolsep}{3.2pt}
\begin{subtable}{\textwidth}
\centering
\begin{tabularx}{\textwidth}{c*{6}{Y}}
\toprule
\multirow{2}{*}{GPU}
& \multirow{2}{*}{\shortstack{Flash\\Attention}}
& \multicolumn{3}{c}{VSA} & \multicolumn{2}{c}{\Sys} \\
\cmidrule(lr){3-5}\cmidrule(lr){6-7}
&  & $s=0.85$ & $s=0.95$ & $s=0.97$
& $s=0.95$ & $s=0.97$ \\
\midrule
RTX 5090 &  7164.56 & 5244.00 & 4627.65 & 4519.46 & 4866.97 & 4381.47 \\
H100     &  2919.98 & 2723.14 & 2399.85 & 2385.24 & 2530.16 & 2168.86 \\
\bottomrule
\end{tabularx}
\caption{480p}
\end{subtable}

\begin{subtable}{\textwidth}
\centering
\begin{tabularx}{\textwidth}{c*{6}{Y}}
\toprule
\multirow{2}{*}{GPU}
& \multirow{2}{*}{\shortstack{Flash\\Attention}}
& \multicolumn{3}{c}{VSA} & \multicolumn{2}{c}{\Sys} \\
\cmidrule(lr){3-5}\cmidrule(lr){6-7}
&  & $s=0.85$ & $s=0.97$ & $s=0.98$
& $s=0.97$ & $s=0.98$ \\
\midrule
RTX 5090 & 24842.18 & 15129.24 & 12661.00 & 11953.88 & 12522.81 & 10626.97 \\
H100     & 9630.09 & 6987.10 &  6209.90 &  6025.72 &  6589.27 &  5706.98 \\
\bottomrule
\end{tabularx}
\caption{720p}
\end{subtable}

\caption{E2E latency (ms) of generating an 81-frame video with 4-step distilled Wan 2.1-1.3B. Sparsity level $s$ is indicated for VSA and \Sys.}
\label{tab:e2e_wan}

\end{table}

\begin{table}[H]
\centering
\setlength{\tabcolsep}{4.2pt}

\begin{tabularx}{\columnwidth}{c*{6}{Y}}
\toprule
& \multicolumn{3}{c}{480p} & \multicolumn{3}{c}{720p} \\
\cmidrule(lr){2-4}\cmidrule(lr){5-7}
\multirow{2}{*}{GPU}
& \multirow{2}{*}{\shortstack{Flash\\Attention}}
& \multicolumn{2}{c}{\Sys}
& \multirow{2}{*}{\shortstack{Flash\\Attention}}
& \multicolumn{2}{c}{\Sys} \\
\cmidrule(lr){3-4}\cmidrule(lr){6-7}
&  & $s=0.95$ & $s=0.97$ &  & $s=0.97$ & $s=0.98$ \\
\midrule
RTX 5090 & 8309.06 &  6094.45 &  5697.59 & --- & --- & --- \\
H100     & 3661.16 &  3228.45 & 2962.28 & 10672.43 & 7634.64 &  6815.19 \\
\bottomrule
\end{tabularx}

\caption{E2E latency (ms) of generating an 81-frame video with Self-Forcing. Sparsity level $s$ is indicated for \Sys. 720p results are excluded on RTX 5090 as the KV cache size causes OOM.}
\label{tab:e2e_sf}

\end{table}

This also translates to significant E2E speedups, as shown in~\Cref{tab:e2e_wan,tab:e2e_sf}. \Sys remains roughly on par with VSA for the same sparsity level. Importantly, \Sys at 95\% sparsity outperforms VSA at 85\% sparsity both in terms of E2E latency (\Cref{tab:e2e_wan}) but also in terms of quality (\Cref{tab:trained_wan}). Focusing on our 480p Self-Forcing results in~\Cref{tab:e2e_sf}, \Sys at 95\% sparsity provides a 36\% E2E speedup on RTX 5090 and a 13\% E2E speedup on H100.

To provide an additional speedup, we apply \texttt{torch.compile} to \Sys and compare it to FA-2 on RTX 5090 for 480p Self-Forcing generation. \textbf{Notably, while FA-2 achieves 11 FPS video generation, \Sys at 95\% sparsity directly achieves 16 FPS} without requiring any additional lossy optimizations such as quantization. 95\% sparsity is the same level at which we demonstrate high quality in~\Cref{sec:quality_eval}. To our knowledge, \Sys is one of the first methods to achieve high-quality generation with auto-regressive models in true real-time on consumer-grade hardware.

%% file: sections/discussion.tex
\section{Related Works}
\textbf{Efficient Attention}. As transformers have widely grown in adoption across a variety of applications, there has been  significant work towards reducing the quadratic cost of softmax attention, both from an algorithmic and from an implementation  perspective. FlashAttention~\citep{dao2022flashattentionfastmemoryefficientexact,shah2024flashattention3fastaccurateattention} is a fused attention kernel implementation that avoids materializing large attention score matrices in GPU memory and instead computes them on the fly in a tiled manner within on-chip SRAM. For video generation, both static~\citep{zhang2025fastvideogenerationsliding,xi2025sparsevideogenacceleratingvideo,li2025radialattentiononlogn} and dynamic~\citep{zhang2025vsafastervideodiffusion,yang2025sparse} sparse attention are explored.

\section{Conclusion}
\label{sec:conclusion}
We have presented \Sys, a principled and efficient attention parameterization for real-time video generation. By analyzing the structure of 3D attention, we showed that its spatiotemporal periodicity and sparse semantic interactions are fundamentally misaligned with existing sparse-attention methods, yet align naturally with the expressive power of Monarch matrices. Building on this insight, our design combines appropriately aligned block structures, the proposed \emph{tiled Monarch parameterization}, and finetuning together with an optimized Triton implementation to deliver substantial speedups while preserving fidelity.

\section*{Acknowledgements}

We gratefully acknowledge access to NVIDIA computing resources. This work was partially supported by Google Research Award, Google ML \& System Junior Faculty Award, Amazon Research Award, Fireworks AI, Intel, Li Auto, Moffett AI, and CMU CyLab Seed funding. This material is also based upon work supported by the National Science Foundation under Grant Nos. CCF-2504353 and CCF-2247014, and by IARPA. Any opinions, findings, conclusions or recommendations expressed are those of the authors and do not necessarily reflect the views of the National Science Foundation.

%% file: sections/supp.tex
\section{MonarchAttention Algorithm}
\label{app:algodetails}
\subsection{Original MonarchAttention}
\label{sec:origmonarchattn}
\textbf{Algorithm 1.} Below are the updates that MonarchAttention uses in a single iteration, derived from optimizing the objective function with respect to $\boldsymbol{L}$ and $\boldsymbol{R}$ individually:
{
\begin{align*}
&\boldsymbol{\alpha}^{(t)}_{R,kjv} = \sum_{\ell} \boldsymbol{L}_{j \ell k}^{(t-1)} \boldsymbol{Q}_{\ell jv} \notag,\ \boldsymbol{c}_{R,kj}^{(t)} = \sum_{\ell} \boldsymbol{L}_{j \ell k}^{(t-1)} \notag \\
&\boldsymbol{\beta}_{R,kji}^{(t)} = \sum_{v} \boldsymbol{\alpha}_{R,kjv}^{(t)} \boldsymbol{K}_{kiv} \notag \\
&\boldsymbol{R}^{(t)} = \softmax_i \left( \boldsymbol{Z}_R^{(t)} \right), \boldsymbol{Z}_{R,kji}^{(t)} = \boldsymbol{\beta}_{R,kji}^{(t)} / \boldsymbol{c}_{R,kj}^{(t)} \notag\\
&\boldsymbol{\alpha}^{(t)}_{L,jkv} = \sum_{i} \boldsymbol{R}_{kji}^{(t)} \boldsymbol{K}_{kiv} \notag,\ \boldsymbol{c}_{L,jk}^{(t)} = \sum_{l} \boldsymbol{R}_{kji}^{(t)} \log \boldsymbol{R}_{kji}^{(t)} \notag\\
&\boldsymbol{\beta}_{L,j \ell k}^{(t)} = \sum_{v} \boldsymbol{\alpha}_{L,j \ell k}^{(t)} \boldsymbol{Q}_{ljv} \notag\\
&\boldsymbol{L}^{(t)} = \softmax_k \left( \boldsymbol{Z}_L^{(t)} \right), \boldsymbol{Z}_{L,j \ell k}^{(t)} = \boldsymbol{\beta}_{L,j \ell k}^{(t)} - \boldsymbol{c}_{L,jk}^{(t)}
\label{eq:monarch_update}
\end{align*}
}
where we interpret $\boldsymbol{Q}$ and $\boldsymbol{K}$ as $b_1 \times b_2 \times d$ tensors. Then, to obtain the output via $\boldsymbol{V}$,
\begin{gather*}
\boldsymbol{Y}_{kjv} = \sum_i \boldsymbol{R}_{kji} \boldsymbol{V}_{kiv},\ \boldsymbol{O}_{\ell jv} = \sum_k \boldsymbol{L}_{j \ell k} \boldsymbol{Y}_{kjv}
\end{gather*}
where we also interpret $\boldsymbol{V}$ as a $b_1 \times b_2 \times d$ tensor. We have also provided pseudocode for the MonarchAttention algorithm in~\Cref{fig:monarch_attn_pseudocode}.

\subsection{Modified MonarchAttention using Tiled Monarch Parameterization}
\label{sec:modifiedmonarchattn}

Under our tiled Monarch parameterization, we adopt the following set of updates for iteratively refining $\boldsymbol{L}$ and $\boldsymbol{R}$, which are also derived from optimizing the objective function with respect to $\boldsymbol{L}$ and $\boldsymbol{R}$ individually:
\begin{align*}
&\boldsymbol{\alpha}^{(t)}_{R, \ell_1 j_1 k_1 i_1 k_2 j_2 v} = \sum_{\ell_2} \boldsymbol{L}_{\ell_1 j_1 k_1 i_1 j_2 \ell_2 k_2}^{(t-1)} \boldsymbol{Q}_{\ell_1 \ell_2 j_1 j_2 v} \\
&\boldsymbol{c}_{R,\ell_1 j_1 k_1 i_1 k_2 j_2}^{(t)} = \sum_{\ell_2} \boldsymbol{L}_{\ell_1 j_1 k_1 i_1 j_2 \ell_2 k_2}^{(t-1)} \\
&\boldsymbol{\beta}_{R, \ell_1 j_1 k_1 i_1 k_2 j_2 i_2}^{(t)} = \sum_{v} \boldsymbol{\alpha}_{R,\ell_1 j_1 k_1 i_1 k_2 j_2 v}^{(t)} \boldsymbol{K}_{k_1 k_2 i_1 i_2 v} \\
&\begin{aligned}[t]
\boldsymbol{R}^{(t)} &= \softmax_{i_2} \left( \boldsymbol{Z}_R^{(t)} \right), \\&\boldsymbol{Z}_{R,\ell_1 j_1 k_1 i_1 k_2 j_2 i_2}^{(t)} = \boldsymbol{\beta}_{R,\ell_1 j_1 k_1 i_1 k_2 j_2 i_2}^{(t)} / \boldsymbol{c}_{R,\ell_1 j_1 k_1 i_1 k_2 j_2}^{(t)}
\end{aligned}\\
&\boldsymbol{\alpha}^{(t)}_{L,\ell_1 j_1 k_1 i_1 j_2 k_2 v} = \sum_{i_2} \boldsymbol{R}_{\ell_1 j_1 k_1 i_1 k_2 j_2 i_2}^{(t)} \boldsymbol{K}_{k_1 k_2 i_1 i_2 v} \\
&\boldsymbol{c}_{L,\ell_1 j_1 k_1 i_1 j_2 k_2}^{(t)} = \sum_{i_2} \boldsymbol{R}_{\ell_1 j_1 k_1 i_1 k_2 j_2 i_2}^{(t)} \log \boldsymbol{R}_{\ell_1 j_1 k_1 i_1 k_2 j_2 i_2}^{(t)} \\
&\boldsymbol{\beta}_{L,\ell_1 j_1 k_1 i_1 j_2 \ell_2 k_2}^{(t)} = \sum_{v} \boldsymbol{\alpha}_{L,\ell_1 j_1 k_1 i_1 j_2 k_2 v}^{(t)} \boldsymbol{Q}_{\ell_1 \ell_2 j_1 j_2 v} \\
&\begin{aligned}[t]
\boldsymbol{L}^{(t)} &= \softmax_{k_1,k_2,i_1} \left( \boldsymbol{Z}_L^{(t)} \right), \\&\boldsymbol{Z}_{L,\ell_1 j_1 k_1 i_1 j_2 \ell_2 k_2}^{(t)} = \boldsymbol{\beta}_{L,\ell_1 j_1 k_1 i_1 j_2 \ell_2 k_2}^{(t)} - \boldsymbol{c}_{L,\ell_1 j_1 k_1 i_1 j_2 k_2}^{(t)}
\end{aligned}\\
\end{align*}
where we interpret $\boldsymbol{Q}$ and $\boldsymbol{K}$ as $c_1 \times \frac{b_1}{c_1} \times c_2 \times \frac{b_2}{c_2} \times d$ tensors. Then to obtain the final attention output via $\boldsymbol{V}$:
\begin{gather*}
\boldsymbol{Y}_{\ell_1 j_1 k_1 i_1 j_2 k_2 v} = \sum_{i_2} \boldsymbol{R}_{\ell_1 j_1 k_1 i_1 k_2 j_2 i_2} \boldsymbol{V}_{k_1 k_2 i_1 i_2 v} \\
\boldsymbol{O}_{\ell_1 \ell_2 j_1 j_2 v} = \sum_{k_1, k_2, i_1} \boldsymbol{L}_{\ell_1 j_1 k_1 i_1 j_2 \ell_2 k_2} \boldsymbol{Y}_{\ell_1 j_1 k_1 i_1 j_2 k_2 v}
\end{gather*}
As these update rules may be difficult to interpret, we have provided pseudocode for this algorithm in~\Cref{fig:tiled_monarch_attn_pseudocode}. In our efficient Triton kernel implementation, we adopt the same approach as MonarchAttention to avoid materializing $\boldsymbol{L}$ and $\boldsymbol{R}$ in HBM and instead use a FlashAttention-like implementation that computes them on-the-fly in SRAM.

\begin{figure}[H]
\centering

\begin{subfigure}[c]{0.43\linewidth}
\centering
\begin{minted}[frame=single,fontsize=\small]{python}
# Q: array of size (N, d)
# K: array of size (N, d)
# V: array of size (N, d)
# T: number of steps
# b1, b2: block sizes

def monarch_attention(Q, K, V, T, b1, b2):
    L = stack(b2 * [eye(b1)])
    Q = Q.view(b1, b2, d)
    K = K.view(b1, b2, d)

    for t in range(T):
        aR = einsum("jkl,ljv->kjv", L, Q)
        bR = einsum("kjv,kiv->kji", aR, K)
        cR = einsum("jkl->kj", L)
        R = softmax(bR / cR[:, :, None],
                    axis=2)

        aL = einsum("kji,kiv->jkv", R, K)
        bL = einsum("jkv,ljv->jkl", aL, Q)
        cL = einsum("kji->jk", R * log(R))
        L = softmax(bL - cL[:, :, None],
                    axis=1)

    V = V.view(b1, b2, d)
    Y = einsum("kji,kiv->jkv", R, V)
    Z = einsum("jkl,jkv->ljv", L, Y)
    O = Z.view(N, d)

    return O
\end{minted}
\caption{MonarchAttention pseudocode}
\label{fig:monarch_attn_pseudocode}
\end{subfigure}
\hfill
\begin{subfigure}[c]{0.56\linewidth}
\centering
\begin{minted}[frame=single,fontsize=\small]{python}
# Q: array of size (N, d)
# K: array of size (N, d)
# V: array of size (N, d)
# T: number of steps
# bb1, bb2: base block sizes
# c1, c2: tiling factors

def tiled_monarch_attention(Q, K, V, T,
                            bb1, bb2, c1, c2):
    ntiles = c1 * c2
    b1 = bb1 // c1
    b2 = bb2 // c2
    L = stack(ntiles * [stack(ntiles * [
                                stack(b2 * [eye(b1)])
                        ])])
    Q = rearrange(Q, "(albj)v -> (ab)ljv",
                    a=c1, b=c2, l=b1, j=b2)
    Q = rearrange(K, "(akbi)v -> (ab)kiv",
                    a=c1, b=c2, k=b1, i=b2)

    for t in range(T):
        aR = einsum("mnjkl,mljv->mnkjv", L, Q)
        bR = einsum("mnkjv,nkiv->mnkji", aR, K)
        cR = einsum("mnjkl->mnkj", L)
        R = softmax(bR / cR[:, :, :, :, None], axis=4)

        aL = einsum("mnkji,nkiv->mnjkv", R, K)
        bL = einsum("mnjkv,mljv->mnjkl", aL, Q)
        cL = einsum("mnkji->mnjk", R * log(R))
        L = softmax(bL - cL[:, :, :, :, None],
                    axis=(1, 3))

    V = rearrange(V, "(akbi)v -> (ab)kiv",
                    a=c1, b=c2, k=b1, i=b2)
    Y = einsum("mnkji,nkiv->mnjkv", R, V)
    Z = einsum("mnjkl,mnjkv->mljv", L, Y)
    O = rearrange(Z, "(ab)ljv -> (albj)v",
                    a=c1, b=c2, l=b1, j=b2)

    return O
\end{minted}
\caption{Tiled MonarchAttention pseudocode}
\label{fig:tiled_monarch_attn_pseudocode}
\end{subfigure}

\caption{Pseudocode for MonarchAttention variants. \textbf{(a)} Standard pseudocode for MonarchAttention, based directly off of the pseudocode provided by~\citet{yaras2025monarchattentionzeroshotconversionfast}. \textbf{(b)} Modified pseudocode for MonarchAttention to support tiled Monarch parameterization.}
\label{fig:all_monarch_pseudocodes}
\end{figure}

\section{Proofs and Analysis}
\subsection{Proof for \Cref{thm:chunk_rank1}}
\label{sec:proofs}

Let $f,\ h,\ w$ denote the number of frames, height, and width respectively, so the total number of tokens is $N = f h w$ and the attention matrix $A \in \mathbb{R}^{N \times N}$ is indexed by pairs of spatiotemporal positions
\[
p = (f_0,\ h_0,\ w_0),\ q = (f_1,\ h_1,\ w_1).
\]
Assume a row-major ordering of tokens along the time, height, and width dimensions. Under the attention map model in \Cref{eq:model}, each entry of $A$ decomposes as
\[
\boldsymbol{A}_{pq}
= \boldsymbol{D}_{pq} + \boldsymbol{S}_{pq} + \epsilon,
\]
where the positional term factorizes along the three axes as
\[
D_{pq}
= d_w(w_0,w_1)d_h(h_0,h_1)d_t(f_0,f_1).
\]
We show that $\boldsymbol{D}$ can be written as a row-wise permutation of a blockwise rank-1
matrix with $\boldsymbol{D}'$ (using permutation $\boldsymbol{P}$).

Assume a standard row-major token ordering on the frame, height, and width indices, the width dimension being contiguous in the flattened token sequence. Concretely, the absolute index $\phi(f_0, h_0, w_0)$ of a token with spatiotemporal indices $(f_0, h_0, w_0)$ is
\[
\phi(f_0, h_0, w_0) = ((f_0 h) + h_0) w + w_0.
\]
We can also consider a permuted token ordering that makes width noncontiguous:
\[
\rho(f_0,h_0,w_0) = w_0 fh + (f_0 h + h_0).
\]
Let $\boldsymbol{P} \in \mathbb{R}^{N\times N}$ be the permutation matrix that maps the token ordering $\rho \to \phi$, i.e.
\[
\boldsymbol{P}_{\phi(f_0,h_0,w_0), \rho(f_0,h_0,w_0)} = 1
\]
and zero elsewhere. Then define
\[
\boldsymbol{D}' = \boldsymbol{P}^\top \boldsymbol{D},
\]
so that $D'$ is obtained from $D$ by a row-wise permutation that maps the original row-major indices from $\phi$ to column-major indices given by $\rho$. Importantly,
\[
\boldsymbol{D} = \boldsymbol{P} \boldsymbol{D}'
\]

We now prove that $\boldsymbol{D}'$ is blockwise rank-1. Using block sizes $(b_1, b_2) = (fh, w)$, we take a 4D blocked view of $\boldsymbol{D}'$ with shape $(b_2, b_1, b_1, b_2)$, so $\boldsymbol{D}'_{j,:,k,:}$ is a block of size $b_1 \times b_2$ for all $j \in [b_2], k \in [b_1]$. Due to the row-wise permutation, the rows of $\boldsymbol{D}'$ (corresponding to query tokens) follow the column-major ordering while the columns (corresponding to key tokens) follow the row-major ordering. For convenience, let us define
\[
\sigma(f_0, h_0) = f_0 h + h_0
\]
to map the frame and height indices for a given token to a combined (row-major) time/height index so that $\sigma (f_0, h_0) \in [b_1]$. Then we can also define $d_{t,h}$:
\[
d_{t,h} ( \sigma(f_0, h_0), \sigma(f_1, h_1)) = d_t (f_0, f_1) d_h (h_0, h_1)
\]
which takes combined time/height indices for two tokens and computes the combined time-wise and height-wise components of the positional attention score for the two tokens.

Let $\boldsymbol{B}^{(j,k)} = \boldsymbol{D}'_{j,:,k,:} \in \mathbb{R}^{b_1 \times b_2}$ denote the $(j,k)$-th block of $\boldsymbol{D}'$. Note that columns of $\boldsymbol{B}^{(j,k)}$ correspond to the key/value tokens with the same time/height index while rows of $\boldsymbol{B}^{(j,k)}$, due to the row-wise permutation for $\boldsymbol{D}'$, correspond to query tokens with the same width index. This means
\[
\boldsymbol{B}^{(j,k)}_{i,l} = d_w(j, l) d_{t,h} (i, k)
\]
Alternatively,
\[
\boldsymbol{B}^{(j,k)} = 
\begin{bmatrix} d_{t,h} (0, k) \\ d_{t,h} (1, k) \\ \vdots \\ d_{t,h} (b_1 - 1, k)  \end{bmatrix} \begin{bmatrix} d_w (j, 0) \\ d_w (j, 1) \\ \vdots \\ d_w (j, b_2 - 1)  \end{bmatrix}^\top
\]
So $\boldsymbol{B}^{(j,k)}$ is rank-1, meaning $\boldsymbol{D}'$ is blockwise rank-1.

Therefore, since \Cref{eq:model} defines
\[
\boldsymbol{A} = \boldsymbol{D} + \boldsymbol{S} + \epsilon
\]
and since
\[
\boldsymbol{D} = \boldsymbol{P} \boldsymbol{D}'
\]
then
\[
\boldsymbol{A} = \boldsymbol{P} \boldsymbol{D}' + \boldsymbol{S} + \epsilon
\]
where $\boldsymbol{P}$ is a permutation matrix, $\boldsymbol{D}'$ is blockwise rank-1 with block sizes $(b_1, b_2)$ that satisfy $b_1 b_2 = fhw$, and $\boldsymbol{S}$ is a sparse matrix.

Note that the proof can be generalized for any of the ``proper" block sizes specified in~\Cref{sec:grouping}, so long as the appropriate sequence flattening order and permutation (defined by $\phi$ and $\rho$) are used.

\subsection{Proof for \Cref{thm:tiled_monarch_strict}}
\label{sec:tiled_monarch_proof}

\textbf{(Containment).}
Let \(\boldsymbol{M}\in\mathcal{M}(b_1,b_2)\) with factors \((\boldsymbol{L},\boldsymbol{R})\). Let \(c_1\mid b_1\), \(c_2\mid b_2\) be tiling factors, and write \(\tilde b_1 := b_1/c_1\), \(\tilde b_2 := b_2/c_2\). Now define tiled factors \((\boldsymbol{L}',\boldsymbol{R}')\) by parameter tying:
for all valid indices,
\[
\boldsymbol{L}'_{\ell_1,j_1,k_1,i_1,j_2,\ell_2,k_2}
:= \boldsymbol{L}_{j,\ell,k}
\quad\text{with}\quad
j=j_1\tilde b_2+j_2,\;\ell=\ell_1\tilde b_1+\ell_2,\;k=k_1\tilde b_1+k_2,
\]
i.e., \(\boldsymbol{L}'\) ignores \(i_1\), and
\[
\boldsymbol{R}'_{\ell_1,j_1,k_1,i_1,\; k_2,j_2,i_2}
:= \boldsymbol{R}_{k,j,i}
\quad\text{with}\quad
i=i_1\tilde b_2+i_2,\;j=j_1\tilde b_2+j_2,\;k=k_1\tilde b_1+k_2,
\]
i.e., \(\boldsymbol{R}'\) ignores \(\ell_1\).
Substituting these definitions into the tiled formula yields
\[
\boldsymbol{L}'_{\ell_1,j_1,k_1,i_1,\; j_2,\ell_2,k_2}\;
\boldsymbol{R}'_{\ell_1,j_1,k_1,i_1,\; k_2,j_2,i_2}
=
\boldsymbol{L}_{j\ell k}\,\boldsymbol{R}_{kji}
=
\boldsymbol{M}_{(\ell b_2+j)(k b_2+i)},
\]
so \(\boldsymbol{M}\in\mathcal{M}_{\mathrm{tile}}(b_1,b_2;c_1,c_2)\). Hence
\(\mathcal{M}(b_1,b_2)\subseteq \mathcal{M}_{\mathrm{tile}}(b_1,b_2;c_1,c_2)\).

\textbf{(Strictness).}
Assume \(c_1>1\) or \(c_2>1\), so at least one of \(\tilde b_1<b_1\) or \(\tilde b_2<b_2\) holds, meaning that
a slice \(\boldsymbol{B}^{(j,k)}\in\mathbb{R}^{b_1\times b_2}\) can contain \emph{multiple} rank-\(1\) tiles.
We construct a tiled matrix whose \((j,k)=(0,0)\) slice has rank \(2\), which is impossible to represent exactly for untiled Monarch.

Consider the slice \(\boldsymbol{B}^{(0,0)}\) (fix \(j=0,k=0\)).
Pick two distinct rows \(r_1\neq r_2\) and two distinct columns \(s_1\neq s_2\) such that
\((r_1,s_1)\) and \((r_2,s_2)\) lie in \emph{different tiles} of the \(c_1\times c_2\) tiling of
\(\{0,\dots,b_1\!-\!1\}\times\{0,\dots,b_2\!-\!1\}\).
(This is always possible when \(c_1>1\) or \(c_2>1\): if \(c_1>1\), choose \(r_1\) and \(r_2\) from different
row-tiles; if \(c_2>1\), choose \(s_1\) and \(s_2\) from different column-tiles.)

Define a tiled Monarch matrix \(\boldsymbol{M}\) by setting all tiles to zero except the two tiles containing
\((r_1,s_1)\) and \((r_2,s_2)\), and within each of these two tiles choose local factors so that the tile equals a
single-entry rank-\(1\) matrix with value \(1\) at that coordinate (and zeros elsewhere). This is feasible because
each tile independently parameterizes an arbitrary rank-\(1\) matrix on its \(\tilde b_1\times\tilde b_2\) support.

Then \(\boldsymbol{B}^{(0,0)}\) has exactly two nonzero entries:
\(\boldsymbol{B}^{(0,0)}_{r_1,s_1}=1\) and \(\boldsymbol{B}^{(0,0)}_{r_2,s_2}=1\).
The \(2\times 2\) submatrix of \(\boldsymbol{B}^{(0,0)}\) restricted to rows \(\{r_1,r_2\}\) and columns
\(\{s_1,s_2\}\) is the identity matrix, hence \(\mathrm{rank}(\boldsymbol{B}^{(0,0)})\ge 2\).
Therefore \(\boldsymbol{M}\notin\mathcal{M}(b_1,b_2)\), since every \((j,k)\) slice of an untiled Monarch matrix
must have rank at most \(1\).
But by construction \(\boldsymbol{M}\in\mathcal{M}_{\mathrm{tile}}(b_1,b_2;c_1,c_2)\).
Thus the containment is strict whenever \(c_1>1\) or \(c_2>1\).

\subsection{Attention Maps as Monarch Matrices}
\label{sec:expressivenessmonarchattention}

In this section, we formalize the notion from~\Cref{sec:monarchfit} that the
attention map, which admits rank-1 blocks based on our case-by-case analysis,
can be represented as a Monarch matrix via permutation. We use the same blocked
indexing and notation as in~\Cref{sec:monarchfit}.

By assumption, the permuted attention matrix
$\widetilde{\boldsymbol{A}} \in \mathbb{R}^{N \times N}$ is partitioned into
blocks $\widetilde{\boldsymbol{A}}_{[j,k]} \in \mathbb{R}^{b_1 \times b_2}$,
with $j \in [b_2], k \in [b_1]$ and $b_1 b_2 = N$, and each block is rank-1.
Thus, for every $(j,k)$ there exist vectors
$\boldsymbol{u}^{(j,k)} \in \mathbb{R}^{b_1}$ and $\boldsymbol{v}^{(j,k)} \in \mathbb{R}^{b_2}$ such that
\[
\widetilde{\boldsymbol{A}}_{[j,k]} = \boldsymbol{u}^{(j,k)} (\boldsymbol{v}^{(j,k)})^\top,
\quad\text{i.e.}\quad
\widetilde{\boldsymbol{A}}_{[j,k]}(\ell,i)
= \boldsymbol{u}^{(j,k)}_{\ell}\, \boldsymbol{v}^{(j,k)}_{i}
\]
We now construct the Monarch factors directly from these blockwise rank-1
decompositions. Define
\[
\boldsymbol{L}_{j\ell k} = \boldsymbol{u}^{(j,k)}_{\ell},
\qquad
\boldsymbol{R}_{kji} = \boldsymbol{v}^{(j,k)}_{i}
\]
Then, under the same blocked indexing used for Monarch matrices
in~\Cref{sec:background}, we have
\[
\widetilde{\boldsymbol{A}}_{[j,k]}(\ell,i)
= \boldsymbol{L}_{j\ell k}\,\boldsymbol{R}_{kji}
\]
Since $\boldsymbol{A} = \boldsymbol{P} \widetilde{\boldsymbol{A}}$, then we also have
\[
\boldsymbol{A}_{ljki} = \boldsymbol{L}_{j\ell k}\,\boldsymbol{R}_{kji}
\]
when we take the 4D view of $\boldsymbol{A}$. This matches the Monarch parameterization exactly. Therefore
$\boldsymbol{A} = \boldsymbol{P} \widetilde{\boldsymbol{A}}$ is a Monarch matrix with block sizes $(b_1,b_2)$ when $\widetilde{\boldsymbol{A}}$ is blockwise rank-1.

\begin{figure*}[h]
  \centering
  \begin{subfigure}{\linewidth}
    \centering
    \includegraphics[width=0.8\linewidth]{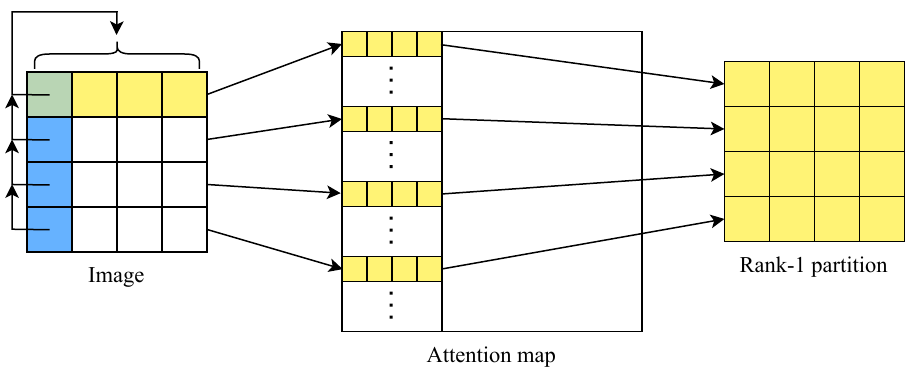}
    \caption{Untiled $(h, w)$ block size partitioning}
    \label{fig:block_partitioning_regular}
  \end{subfigure}
  \hfill
  \begin{subfigure}{\linewidth}
    \centering
    \includegraphics[width=0.8\linewidth]{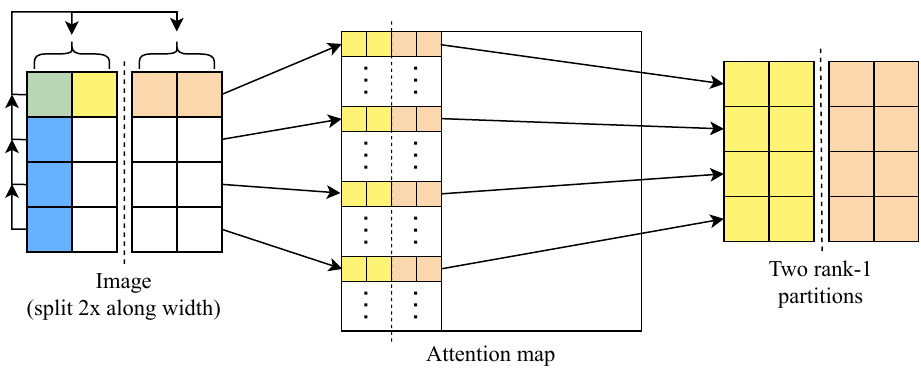}
    \caption{Tiled $(h, \frac{w}{2})$}
    \label{fig:block_partitioning_tiled}
  \end{subfigure}
  \caption{Illustration of untiled/tiled block size partitioning schemes.}
  \label{fig:block_partitioning}
\end{figure*}

\subsection{Block Size Analysis}
\label{sec:blocksizeanalysis}

For simplicity, let us consider the single image case ($f=1$). Based on the block size analysis conducted in Section \cref{sec:grouping}, block sizes $(h, w)$ would enable a proper Monarch parameterization of the positional component of the $hw \times hw$ attention map. With these block sizes, the parameterization effectively forms partitions of the attention map that contain pairwise attention scores between all tokens in a given column and all tokens in a given row of the image. This allows each partition to be rank-1 (under the assumptions from~\Cref{eq:model}), enabling a proper parameterization. We illustrate this blocking scheme in \cref{fig:block_partitioning_regular}. Using different block sizes (e.g. $(2h, \frac{w}{2})$) would cause each partition to span all tokens along multiple columns or multiple rows of the image, which would remove the rank-1 structure as previously noted. Importantly, this restricts block sizes to be equivalent to the video dimensions.

In ~\cref{sec:tiled}, we introduce a tiled Monarch parameterization to reduce this constraint. Effectively, our modified parameterization to an extent enables decoupling the block size selection from the actual video dimensions. We illustrate how this parameterization functions in~\Cref{fig:block_partitioning_tiled} in the case where we select one of the block sizes as $\frac{w}{2}$. In the tiled parameterization, by applying the appropriate permutation/blocking, we gain much more flexibility in how we select our block sizes, which decide how rank-1 partitions are formed.

\section{Extended Results}

We provide the extended VBench results corresponding to our previous evaluations in~\Cref{sec:quality_eval,sec:training_free_ablation}. The extended results generally corroborate the conclusions we drew from our overall empirical evaluations.

\subsection{Extended Trained Results}

\begin{table}[H]
\centering
\small
\renewcommand{\arraystretch}{1.2}
\begin{adjustbox}{max width=\textwidth}
\begin{tabular}{lcc}
\toprule
Metric & Dense Attention & \Sys (95\% sparse) \\
\midrule
Subject Consistency    & 0.959 & 0.954 \\
Background Consistency & 0.960 & 0.956 \\
Temporal Flickering    & 0.990 & 0.991 \\
Motion Smoothness      & 0.985 & 0.984 \\
Dynamic Degree         & 0.653 & 0.747 \\
Aesthetic Quality      & 0.650 & 0.640 \\
Imaging Quality        & 0.680 & 0.672 \\
Object Class           & 0.946 & 0.949 \\
Multiple Objects       & 0.859 & 0.855 \\
Human Action           & 0.966 & 0.968 \\
Color                  & 0.872 & 0.884 \\
Spatial Relationship   & 0.776 & 0.776 \\
Scene                  & 0.580 & 0.575 \\
Appearance Style       & 0.205 & 0.202 \\
Temporal Style         & 0.244 & 0.245 \\
Overall Consistency    & 0.265 & 0.266 \\
\midrule
Quality Score          & 0.844 & 0.846 \\
Semantic Score         & 0.804 & 0.805 \\
Total Score            & 0.836 & 0.838 \\
\bottomrule
\end{tabular}
\end{adjustbox}
\caption{Extended VBench scores from~\Cref{tab:trained_self_forcing} for base model and trained \Sys on Self-Forcing.}
\label{tab:trained_self_forcing_extended}
\end{table}

\begin{table}[H]
\centering
\scriptsize
\renewcommand{\arraystretch}{1.2}
\begin{adjustbox}{max width=\width}
\begin{tabular}{@{} l c c c c c c @{}}
\toprule
& \multicolumn{3}{c}{\textbf{4-step}} & \multicolumn{3}{c}{\textbf{50-step}} \\
\cmidrule(lr){2-4}\cmidrule(lr){5-7}
Metric
& \shortstack{Dense\\Attention}
& \shortstack{VSA\\(90\% sparse)}
& \shortstack{\Sys\\(95\% sparse)}
& \shortstack{Dense\\Attention}
& \shortstack{VSA\\(90\% sparse)}
& \shortstack{\Sys\\(95\% sparse)} \\
\midrule
Subject Consistency    & 0.955 & 0.975 & 0.952 & 0.959 & 0.927 & 0.955 \\
Background Consistency & 0.949 & 0.948 & 0.957 & 0.975 & 0.973 & 0.974 \\
Temporal Flickering    & 0.982 & 0.979 & 0.991 & 0.996 & 0.996 & 0.996 \\
Motion Smoothness      & 0.984 & 0.977 & 0.985 & 0.984 & 0.979 & 0.984 \\
Dynamic Degree         & 0.775 & 0.803 & 0.744 & 0.633 & 0.619 & 0.597 \\
Aesthetic Quality      & 0.644 & 0.576 & 0.624 & 0.656 & 0.644 & 0.653 \\
Imaging Quality        & 0.684 & 0.634 & 0.664 & 0.663 & 0.619 & 0.659 \\
Object Class           & 0.966 & 0.957 & 0.950 & 0.946 & 0.921 & 0.930 \\
Multiple Objects       & 0.872 & 0.814 & 0.850 & 0.851 & 0.791 & 0.862 \\
Human Action           & 0.966 & 0.910 & 0.934 & 0.960 & 0.966 & 0.972 \\
Color                  & 0.853 & 0.930 & 0.874 & 0.883 & 0.872 & 0.879 \\
Spatial Relationship   & 0.774 & 0.798 & 0.739 & 0.779 & 0.735 & 0.795 \\
Scene                  & 0.548 & 0.509 & 0.563 & 0.568 & 0.515 & 0.569 \\
Appearance Style       & 0.200 & 0.220 & 0.199 & 0.217 & 0.218 & 0.217 \\
Temporal Style         & 0.244 & 0.234 & 0.236 & 0.248 & 0.243 & 0.248 \\
Overall Consistency    & 0.266 & 0.254 & 0.261 & 0.270 & 0.264 & 0.269 \\
\midrule
Quality Score          & 0.846 & 0.828 & 0.843 & 0.846 & 0.827 & 0.841 \\
Semantic Score         & 0.800 & 0.793 & 0.788 & 0.810 & 0.785 & 0.812 \\
Total Score            & 0.837 & 0.821 & 0.832 & 0.839 & 0.819 & 0.835 \\
\bottomrule
\end{tabular}
\end{adjustbox}
\caption{Extended VBench scores from~\Cref{tab:trained_wan} for Wan2.1-1.3B (base 50-step and distilled 4-step models) for dense attention as well as trained VSA and \Sys.}
\label{tab:trained_wan_extended}
\end{table}

\begin{table}[H]
\centering
\scriptsize
\resizebox{\textwidth}{!}{%
\begin{tabular}{@{}lccccc@{}}
\toprule
Metric 
& \shortstack{Dense\\Attention} 
& \shortstack{SVG\\(85\% sparse)} 
& \shortstack{RadialAttention\\(85\% sparse)} 
& \shortstack{Exact top-$k$\\(85\% sparse)} 
& \shortstack{\Sys\\(90\% sparse)} \\
\midrule
Subject Consistency    & 0.959 & 0.855 & 0.970 & 0.942 & 0.949 \\
Background Consistency & 0.960 & 0.935 & 0.963 & 0.935 & 0.954 \\
Temporal Flickering    & 0.990 & 0.972 & 0.993 & 0.985 & 0.986 \\
Motion Smoothness      & 0.985 & 0.976 & 0.988 & 0.981 & 0.983 \\
Dynamic Degree         & 0.653 & 0.097 & 0.539 & 0.806 & 0.753 \\
Aesthetic Quality      & 0.650 & 0.398 & 0.632 & 0.595 & 0.643 \\
Imaging Quality        & 0.680 & 0.608 & 0.701 & 0.675 & 0.698 \\
Object Class           & 0.946 & 0.026 & 0.846 & 0.809 & 0.963 \\
Multiple Objects       & 0.859 & 0.004 & 0.718 & 0.558 & 0.878 \\
Human Action           & 0.966 & 0.084 & 0.808 & 0.748 & 0.960 \\
Color                  & 0.872 & 0.848 & 0.897 & 0.916 & 0.880 \\
Spatial Relationship   & 0.776 & 0.022 & 0.817 & 0.683 & 0.829 \\
Scene                  & 0.580 & 0.000 & 0.319 & 0.213 & 0.563 \\
Appearance Style       & 0.205 & 0.206 & 0.192 & 0.199 & 0.196 \\
Temporal Style         & 0.244 & 0.035 & 0.235 & 0.224 & 0.241 \\
Overall Consistency    & 0.265 & 0.047 & 0.244 & 0.234 & 0.264 \\
\midrule
Quality Score          & 0.844 & 0.715 & 0.841 & 0.834 & 0.847 \\
Semantic Score         & 0.804 & 0.214 & 0.718 & 0.658 & 0.808 \\
Total Score            & 0.836 & 0.615 & 0.816 & 0.799 & 0.839 \\
\bottomrule
\end{tabular}%
}
\caption{Extended VBench scores from~\Cref{tab:training_free_self_forcing} for Self-Forcing, comparing dense attention, several sparse baselines, and \Sys all in a training-free setting. For RadialAttention and SVG, the first denoising step (out of 4 for Self-Forcing) and first attention block are retained as dense attention.}
\label{tab:tf_sf_extended}
\end{table}

\begin{table}[H]
\centering
\scriptsize
\begin{tabular}{lcccccc}
\toprule
Metric 
& \shortstack{Dense\\Attention} 
& \shortstack{SVG\\(85\% sparse)}
& \shortstack{SVG2\\(85\% sparse)}
& \shortstack{SVG2\\(90\% sparse)}
& \shortstack{RadialAttention\\(85\% sparse)}
& \shortstack{\Sys\\(90\% sparse)}\\
\midrule
Subject Consistency    & 0.955 & 0.901 & 0.946 & 0.934 & 0.832 & 0.952 \\
Background Consistency & 0.949 & 0.949 & 0.949 & 0.945 & 0.881 & 0.960 \\
Temporal Flickering    & 0.982 & 0.976 & 0.986 & 0.985 & 0.979 & 0.991 \\
Motion Smoothness      & 0.984 & 0.972 & 0.986 & 0.984 & 0.982 & 0.988 \\
Dynamic Degree         & 0.775 & 0.853 & 0.742 & 0.742 & 0.361 & 0.653 \\
Aesthetic Quality      & 0.644 & 0.526 & 0.623 & 0.598 & 0.539 & 0.622 \\
Imaging Quality        & 0.684 & 0.532 & 0.654 & 0.625 & 0.545 & 0.686 \\
Object Class           & 0.966 & 0.524 & 0.921 & 0.896 & 0.757 & 0.956 \\
Multiple Objects       & 0.872 & 0.357 & 0.741 & 0.655 & 0.534 & 0.884 \\
Human Action           & 0.966 & 0.900 & 0.952 & 0.928 & 0.954 & 0.964 \\
Color                  & 0.853 & 0.733 & 0.860 & 0.865 & 0.805 & 0.893 \\
Spatial Relationship   & 0.774 & 0.376 & 0.669 & 0.654 & 0.463 & 0.821 \\
Scene                  & 0.548 & 0.231 & 0.537 & 0.510 & 0.408 & 0.549 \\
Appearance Style       & 0.200 & 0.213 & 0.201 & 0.205 & 0.222 & 0.198 \\
Temporal Style         & 0.244 & 0.202 & 0.238 & 0.234 & 0.236 & 0.240 \\
Overall Consistency    & 0.266 & 0.218 & 0.264 & 0.260 & 0.252 & 0.263 \\
\midrule
Quality Score          & 0.846 & 0.792 & 0.837 & 0.824 & 0.738 & 0.841 \\
Semantic Score         & 0.800 & 0.563 & 0.765 & 0.743 & 0.681 & 0.807 \\
Total Score            & 0.837 & 0.746 & 0.823 & 0.808 & 0.727 & 0.834 \\
\bottomrule
\end{tabular}
\caption{Extended VBench scores from~\Cref{tab:training_free_wan} for 4-step distilled Wan 2.1-1.3B, comparing dense attention, several sparse baselines, and \Sys all in a training-free setting.}
\label{tab:tf_wan_extended}
\end{table}
\clearpage

\subsection{Example Generations}

\begin{figure*}[t]
  \centering
  \includegraphics[width=\linewidth]{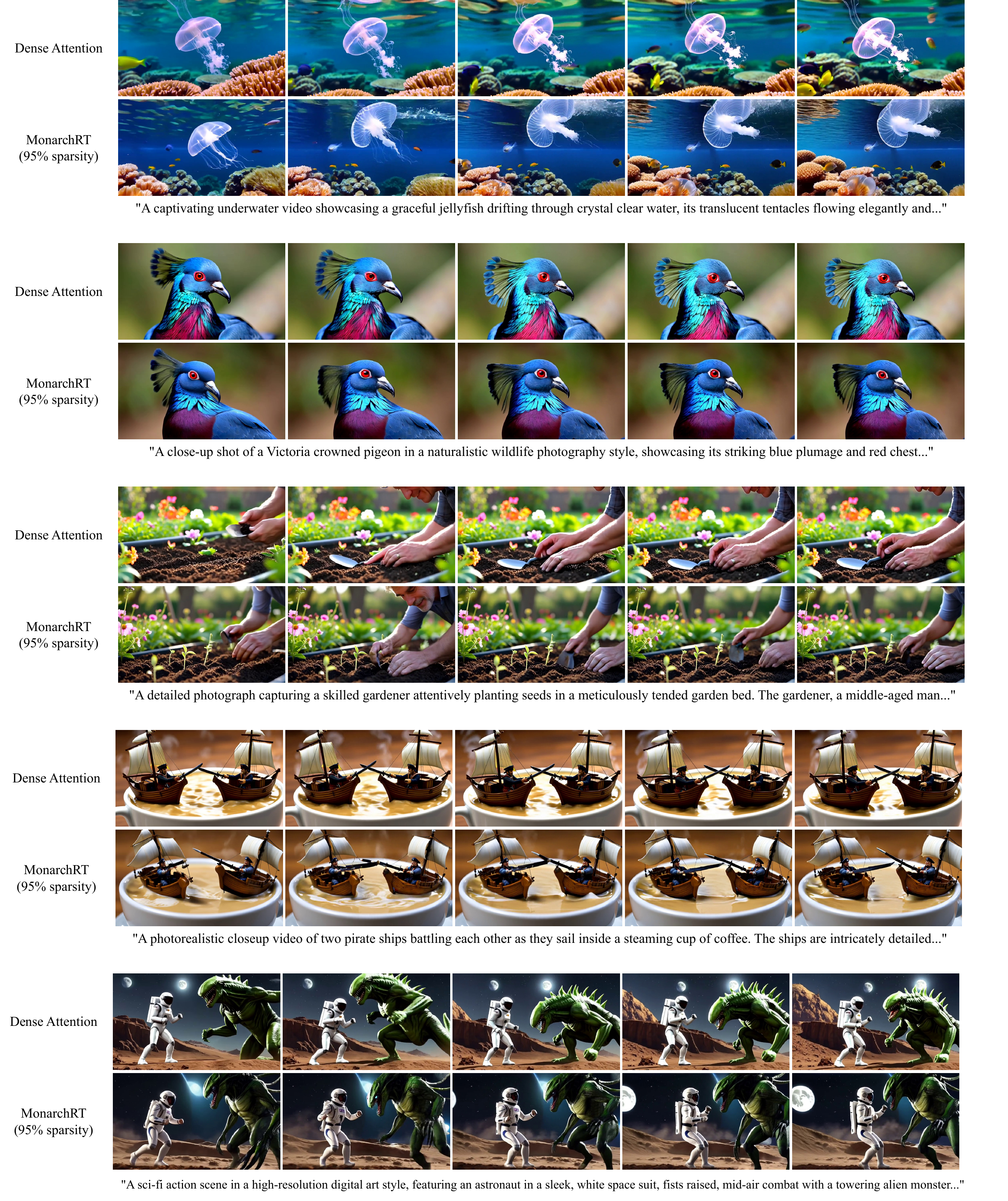}
  \caption{Example generations for dense baseline and \Sys (with 95\% sparsity) on the same prompts for Self-Forcing.}
  \label{fig:sf_samples}
\end{figure*}

\begin{figure*}[t]
  \centering
  \includegraphics[width=\linewidth]{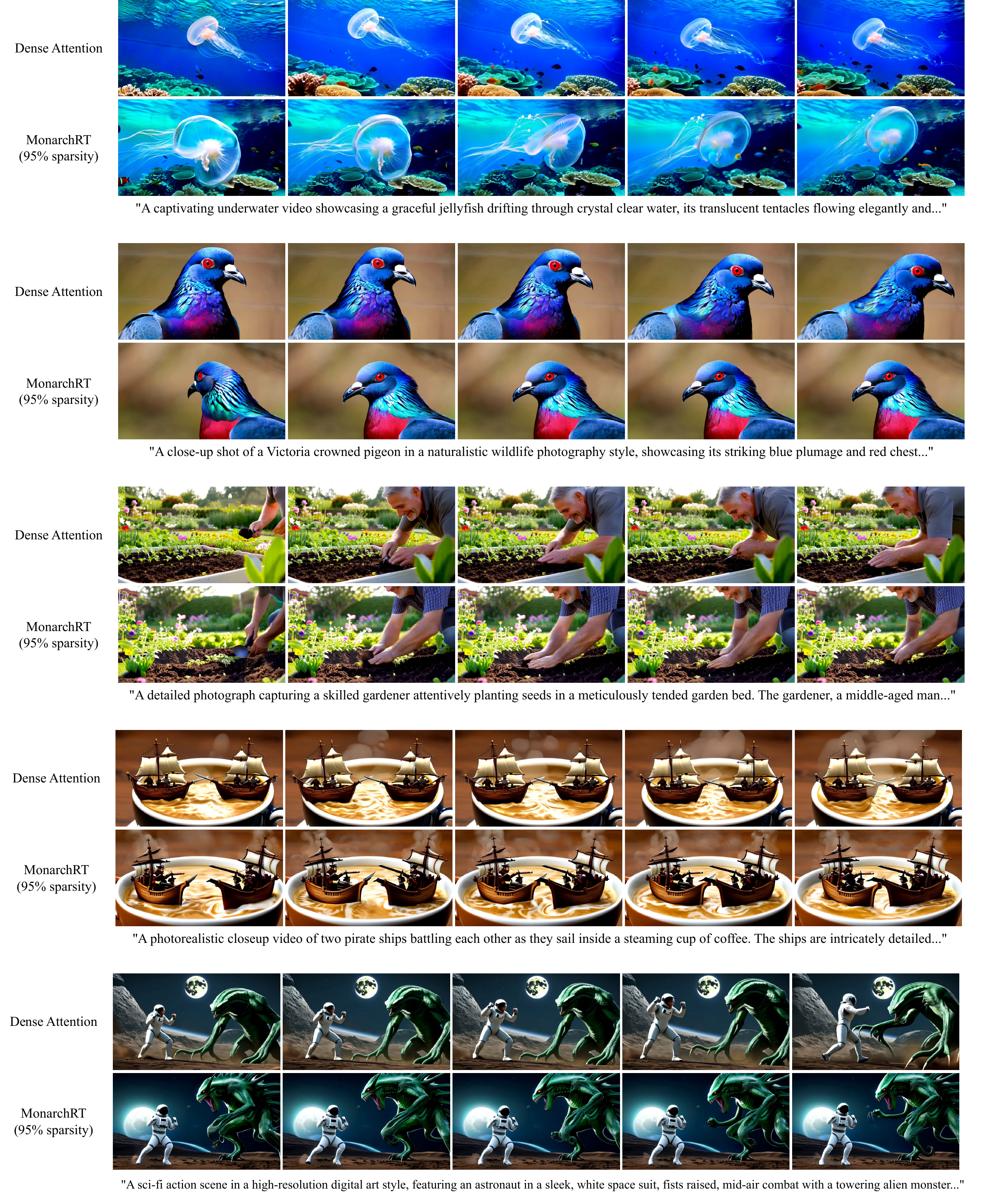}
  \caption{Example generations for dense baseline and \Sys (with 95\% sparsity) on the same prompts for Wan 2.1-1.3B (50-step bidirectional model).}
  \label{fig:wan_samples}
\end{figure*}

We provide additional example generations for Self-Forcing and Wan in Figures \ref{fig:sf_samples} and \ref{fig:wan_samples} respectively. We use 5 prompts from MovieBench~\citep{polyak2024movie} to generate videos for the dense model and \Sys. The sample videos demonstrate that \Sys can achieve comparable visual quality to the dense model.

\begin{figure}[H]
  \centering
  \includegraphics[width=\linewidth]{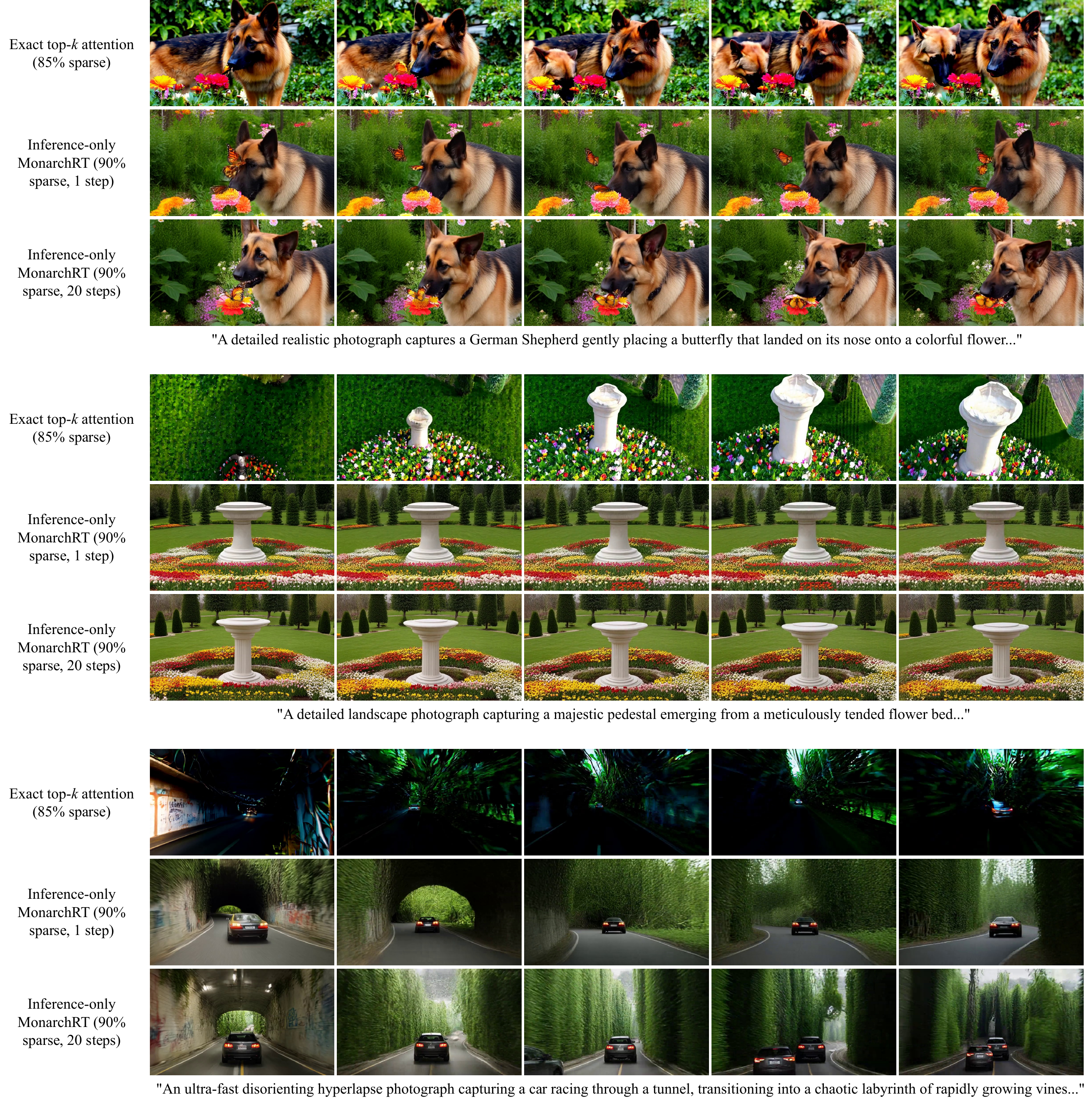}
  \caption{Example generations on Self-Forcing for exact top-$k$ and \Sys with 1 and 20 iterative refinement steps (training-free).}
  \label{fig:sf_inference_examples}
\end{figure}

